\documentclass[journal,10pt]{IEEEtran}
\usepackage{amsmath,graphicx}
\usepackage[ruled,vlined]{algorithm2e}
\usepackage{lineno}
\usepackage{amssymb}
\usepackage{cite}
\usepackage{soul,color,booktabs}
\usepackage{cleveref}

\newcommand{\EQ}{\begin{equation}}
\newcommand{\NQ}{\end{equation}}
\newcommand{\ER}{\begin{eqnarray}}
\newcommand{\NR}{\end{eqnarray}}
\newcommand{\ERS}{\begin{eqnarray*}}
\newcommand{\NRS}{\end{eqnarray*}}
\newcommand{\bit}{\begin{itemize}}
\newcommand{\ben}{\begin{enumerate}}
\newcommand{\eben}{\end{enumerate}}
\newcommand{\ebit}{\end{itemize}}

\newcommand{\bbf}{{\bf f}}

\newcommand{\bp}{{\bf p}}

\newcommand{\bx}{{\bf x}}

\newcommand{\bV}{{\bf V}}

\ifCLASSINFOpdf
\else
\fi
\begin{document}
%
\title{A Two-Stage Approach to Few-Shot Learning for Image Recognition}
%
%
%

\author{Debasmit~Das and~C.S. George~Lee
\thanks{Debasmit Das (dsdas@purdue.edu) and C. S. George Lee (csglee@purdue.edu) are with the School of Electrical and Computer Engineering, Purdue University, West Lafayette, IN, 47907 USA} }

%
%

\markboth{Journal of \LaTeX\ Class Files,~Vol.~14, No.~8, August~2015}%
{Shell \MakeLowercase{\textit{et al.}}: Bare Demo of IEEEtran.cls for IEEE Journals}
%



\maketitle

\begin{abstract}
This paper proposes a multi-layer neural network structure 
for few-shot image recognition of novel categories. 
The proposed multi-layer neural network architecture encodes transferable knowledge 
extracted from a large annotated dataset of base categories. 
This architecture is then applied to novel categories containing only a few samples. 
The transfer of knowledge is carried out at the feature-extraction and the classification levels 
distributed across the two training stages. 
In the first-training stage, we introduce the relative feature to capture the structure of the data 
as well as obtain a low-dimensional discriminative space. 
Secondly, we account for the variable variance of different categories 
by using a network to predict the variance of each class. 
Classification is then performed by computing the Mahalanobis distance 
to the mean-class representation in contrast to previous approaches that used the Euclidean distance. 
In the second-training stage, 
a category-agnostic mapping is learned from the mean-sample representation 
to its corresponding class-prototype representation. This is because the mean-sample representation may not accurately represent the novel category prototype.  
Finally, we evaluate the proposed network structure 
on {four} standard few-shot image recognition datasets, 
where our proposed few-shot learning system produces competitive performance compared to previous work. 
We also extensively studied and analyzed the contribution of each component of our proposed framework.
\end{abstract}

\footnotetext[2]{This work was supported in part by the National
Science Foundation under Grant IIS-1813935. 
Any opinion, findings,
and conclusions or recommendations expressed in this material are
those of the authors and do not necessarily reflect the views of
the National Science Foundation. }
\footnotetext[2]{We also gratefully acknowledge the support of NVIDIA Corporation 
for the donation of a TITAN XP GPU used for this research.}


%
\IEEEpeerreviewmaketitle

\section{Introduction}
\begin{figure}[tbp]
\centering
\includegraphics[width=6cm]{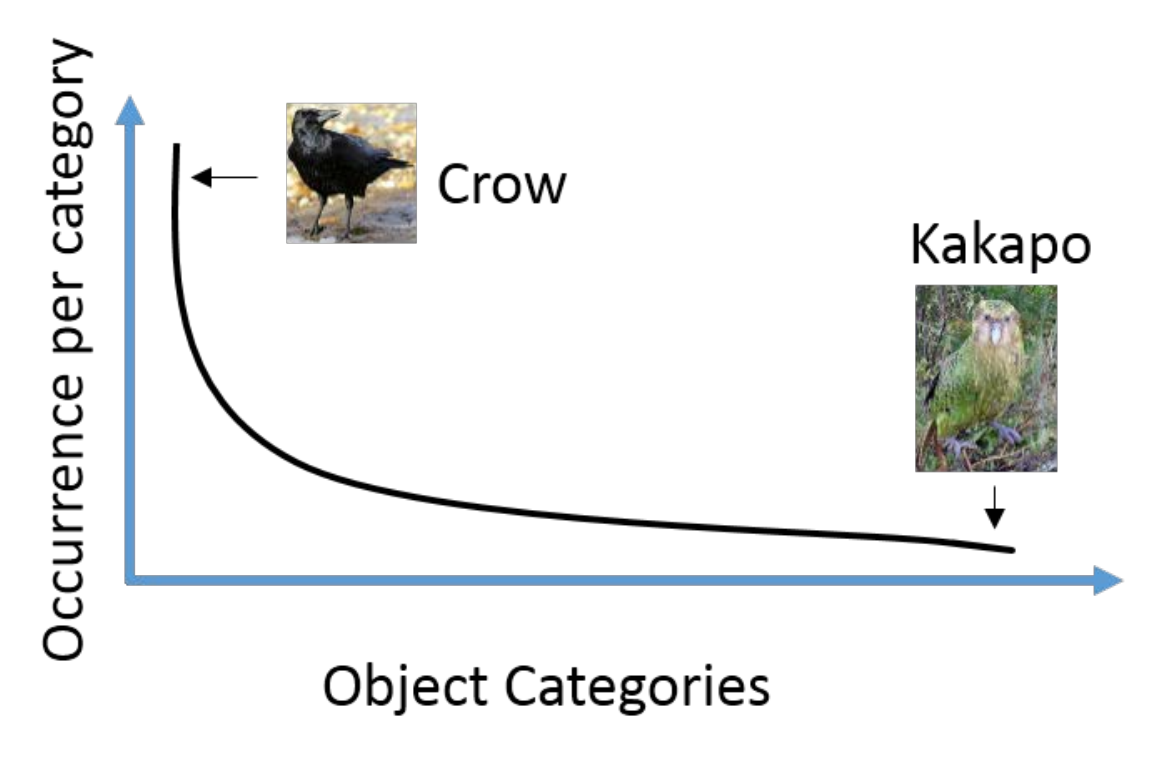}
\caption{Object categories follow a long tailed distribution with a lot of rare classes and very few common classes.}
\label{fig:long-tail}
\vspace*{-0.2in}
\end{figure}

For the past decade, deep convolutional neural networks (CNN) 
have produced excellent results in visual recognition tasks 
such as object recognition, scene classification, etc.~\cite{deng2009imagenet,krizhevsky2012imagenet,zhou2014learning}. 
A CNN learns to recognize a large quantity of visual categories 
by training on a large collection of annotated images 
using a  gradient-descent technique~\cite{lecun1998gradient}.
Although the training procedure is computationally intensive, 
it can be parallelized using a Graphics Processing Unit (GPU). 
Even after a long training period, 
the CNN can only recognize a fixed set of image categories. 
To learn to recognize novel categories, 
one has to collect new training data and re-train 
the CNN model with further adjustments. 
Unfortunately, in some cases, 
there might not be enough labeled data available for training a novel category. 
This results in a long-tailed distribution of object categories~\cite{zhu2014capturing} as shown in Fig.~\ref{fig:long-tail}. In such a long-tailed distribution, only a few object categories occur frequently. 
{Thus}, we can obtain lot of samples from these categories. However, there are lots of categories which occur very rarely. For these object categories, we can obtain only a very few samples. As an example, a crow is a bird {that} we see very often. Therefore we can collect lot of crow samples 
with sufficient variability. On the other hand,  samples of a rare bird kakapo are very difficult to obtain.  

Research on learning novel categories from 
a few samples is termed {\em few-shot learning}. 
Most previous methods tackle few-shot learning by assuming 
access to a large labeled training database as base categories. 
Using this large database, the goal of few-shot image recognition systems 
is to recognize any novel category accurately from just a few samples of that category. 

Traditional supervised learning using a few samples for training 
often causes overfitting and results in poor generalization. 
The poor performance in generalization is due to the following reasons: 
Firstly, it is related to the fundamental problem known as the curse of dimensionality. 
The sparsity of the feature volume due to less number of samples 
in such a high-dimensional image feature space aggravates the problem of overfitting. 
Secondly, the use of only a few training samples would not be able to represent the overall variation of a class. 
Hence, the true spread of the class distribution remains unknown 
and the classification boundaries are poorly estimated. 
Also, the few training samples of a class might be sampled 
near the edge of the class distribution.  
As a result, the mean of these training samples would not be close to the true mean of the class. Therefore, the mean would not accurately represent 
the location of the class in the feature space, resulting in mis-classification. 

In this paper, 
we propose solutions to each of the above problems.
Firstly,  to address the problem of high-dimensionality, 
we propose a low-dimensional discriminative space called the relative-feature space. 
In this space, the relative feature of a sample is represented as 
a vector of distances between the training samples in a training batch. 
Since the number of training samples is less, 
the dimensionality of this relative feature space 
will be a lot less than the dimensionality of the original absolute feature space. 
Also, the features will be discriminative since instances 
from the same classes are expected to cluster 
and would have similar pairwise inter-class and intra-class distances. 
Additional benefit of using these relative features is that 
they extract second-order structural information about the dataset to assist recognition. 
Using higher-order features beyond the second-order relative features would not have the added benefit of having a low-dimensional feature space. Therefore, the combination of relative features and absolute features presents better performance in recognition.
Secondly, to address the uncertain variance of categories, 
we propose a trainable neural network (NN) 
as a module to predict the variance of each category.  
Finally, we propose to learn a category-agnostic transformation 
from the class-mean representation to the class-prototype representation. 
As a result, more accurate locations of the class can be obtained from 
the mean of a few samples. 

The contributions of this paper are both at the feature-extraction 
and classification stage of the few-shot object recognition system.
They can be summarized as follows: 
(a) A novel relative-feature descriptor in combination with 
the original absolute deep-feature descriptor for object recognition,
(b) A framework for learning class variances 
in order to compute the Mahalanobis distances to class prototypes, 
(c) Additional training pipeline in order to learn a category-agnostic transformation 
from the class-mean representation to the class prototype.
The training of the two stages has not been carried out jointly 
since the category-agnostic transformation assumes that a robust representation 
has already been learned for the images.
Finally, we have conducted extensive experimentation and analyses 
on four standard datasets to verify the validity of 
the proposed two-stage few-shot learning framework for image recognition.

This paper is organized into five sections.
Section II discusses related work 
and Section III describes our proposed approach. 
Section IV provides experimental results and discussion. 
This is followed by conclusion and future direction in Section V.

\section{Related Work}
The field of few-shot learning has shown increased interest in the past decade. Most of the earlier methods used a Bayesian approach 
of introducing priors to facilitate the few-shot learning. 
Li et al.~\cite{fei2006one} used a global prior 
while Salakhutdinov et al.~\cite{1shotsala} used a super-category-level prior. 
For application-specific tasks like handwriting recognition, 
generative models have been proposed that can produce characters 
from parts~\cite{wong2015one} or strokes~\cite{lake2013one}. 
For object recognition, a hierarchical Bayesian program has been proposed
to utilize compositional and causal approaches to create a probabilistic generative model 
for visual objects~\cite{lake2015human,lake2011one}. 
Some ad-hoc approaches to address few-shot learning were to carry out data augmentation 
by harnessing unlabeled data~\cite{chapelle2009semi}, 
by transformation and adding noise~\cite{chatfield2014return, dosovitskiy2014discriminative}, 
and by synthesizing artificial examples~\cite{goodfellow2014generative,
hariharan2017low,WangCVPR2018a,mehrotra2017generative} 
or using compositional representations~\cite{zhu2015we, denton2015deep}. {More recent methods that used generative modeling include the auto-encoder~\cite{schwartz2018delta} and variations of adversarial-network-based architectures~\cite{zhang2018metagan,gao2018low}.} 
However, most of these generative methods require lots of efforts to generate data, 
otherwise the generated data do not represent the actual data distribution properly. 
Thus, recent methods mostly take a metric-learning or a meta-learning approach to few-shot learning.

Metric learning approaches strive to preserve class neighborhood structure; 
that is, the representations are learned such that 
features from the same class are clustered together 
while features from different classes are kept far apart. 
As a result, novel-class features are expected to have more room 
for classification error. 
Koch et al.~\cite{koch2015siamese} used Siamese Networks 
to match a training example of a novel category to a test example. 
The training was carried out using an object recognition dataset. 
Vinyals et al.~\cite{vinyals2016matching} proposed Matching Networks, 
which used a nearest-neighbor classifier in addition to 
an attention mechanism over the training samples. 
Prototypical Networks~\cite{snell2017prototypical} extended 
nearest-mean classifiers~\cite{mensink2013distance} 
and learned to classify query samples by computing Euclidean distances 
to prototype features. 
As an extension to Prototypical Networks, 
Sung et al.~\cite{sung2018learning} learned a distance metric 
instead of using a predefined distance function. 
{A more recent method~\cite{oreshkin2018tadam} 
used a metric learning approach, 
where the metric is scaled and adapted {based} on the task.}   

On the other hand, meta-learning methods for few-shot learning use a learning-to-learn scheme, where a model extracts useful transferable knowledge about the learning procedure from a large collection of tasks. This helps in quickly learning the novel task which, in our case, is image recognition for novel categories. 
Ravi and Larochelle~\cite{ravi2016optimization} 
used {Long-Short-Term Memory (LSTM)}~\cite{hochreiter1997long} 
to train a meta-learner to produce model parameter updates 
for optimization of a base learner on a task. 
This method basically learns the optimization procedure 
using data from a number of auxiliary tasks. 
The work on learning-to-learn~\cite{andrychowicz2016learning} approach 
to few-shot learning is also closely related to the {learning-to-optimize} technique. 
Finn et al.~\cite{finn2017model} built upon this work to focus on 
learning the initial parameter for gradient descent so that 
{the learner can be optimized for a new task in a few iterations.}
Mishra et al.~\cite{mishra2017simple} introduced temporal convolutions 
to predict the label of a test example, 
given a sequence of labeled samples and the unlabeled test sample. 
{The transductive propagation network~\cite{liu2018learning} 
classifies the whole test dataset using a graph-based label propagation mechanism. 
They use an end-to-end meta-learning framework 
to learn the feature embedding and graph construction simultaneously. 
Sun et al.~\cite{Sun_2019_CVPR} used a meta-transfer learning mechanism 
that shifts and scales neural network weights for new tasks. 
Similarly, Munkhdalai et al.~\cite{munkhdalai2017rapid} 
proposed a meta-learning scheme that shifts the neuron activations 
depending on task-specific parameters.}

Alternatively, few-shot learning methods include memory-based {models~\cite{santoro2016meta,munkhdalai2017meta,ramalho2019adaptive}} 
that store selective relevant information and use that for comparison at test time. 
Attentive comparators~\cite{shyam2017attentive} 
compare patches of images sequentially through an attention mechanism 
and then arrive at a prediction. 
{Qiao et al.~\cite{qiao2018few} learned 
a category-agnostic mapping from activations to parameters 
that allowed fast generalization to novel categories. 
A similar idea~\cite{qi2018low} was used to imprint weights 
for the classification layer of the novel categories. 
Bertinetto et al.~\cite{bertinetto2018meta} used 
a differential closed-form solver based on ridge regression 
for fast adaptation to novel categories. 
Some methods extended existing machine learning concepts 
like graph neural networks~\cite{garcia2017few} and 
information retrieval~\cite{triantafillou2017few} 
to few-shot learning.} 
For a more comprehensive survey on few-shot learning, 
one can refer to~\cite{sslsurvey,fslsurvey}.
\begin{figure*}[ht]
\centering
\includegraphics[width=18cm]{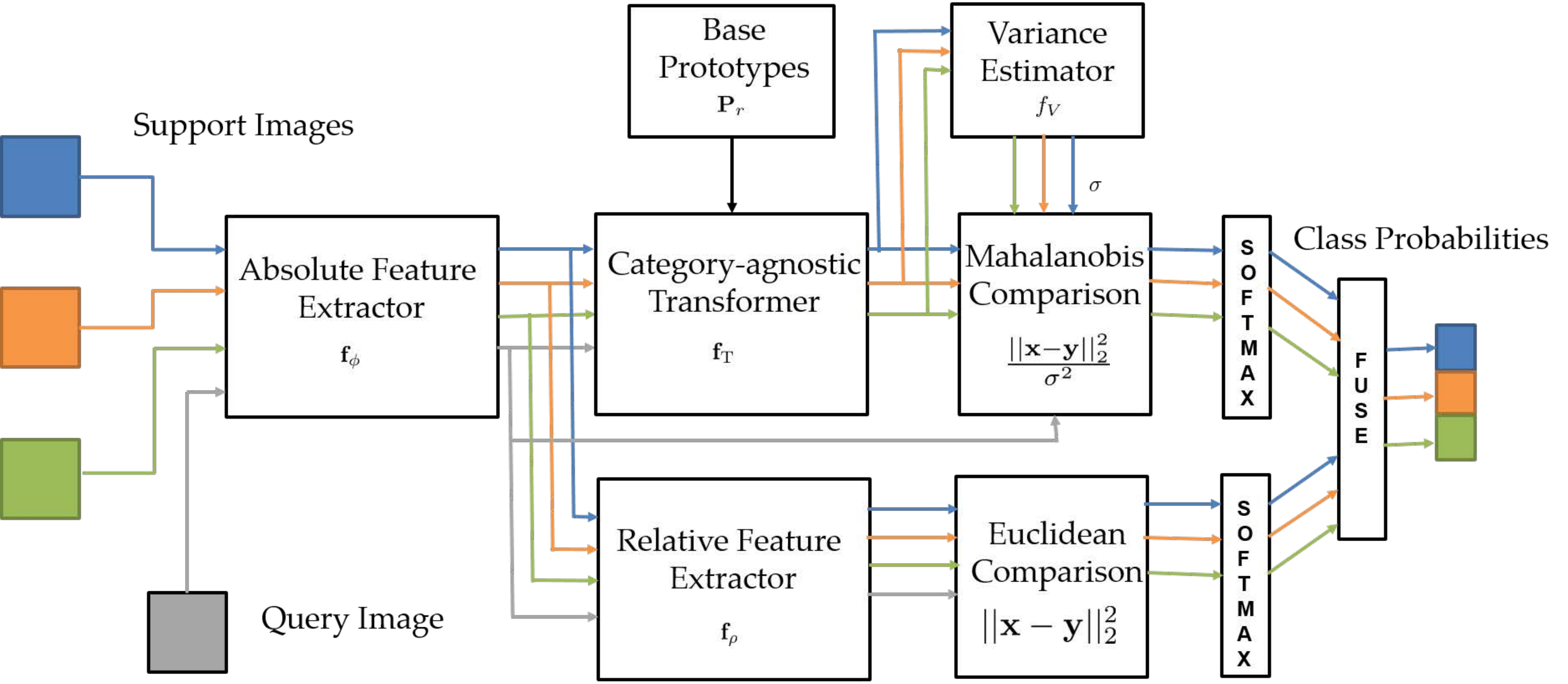}
\caption{Overall framework for the proposed approach for a 3-way 1-shot inference scenario. A single image from each of the 3 classes (classes are shown in different colors) are used as support examples while a single query image is used. The output is the probability of the query example belonging to each of the 3 classes.}
\label{fig:whole}
\end{figure*}

\section{Proposed Approach}

\subsection{Problem Definition and Formulation}
Our proposed few-shot learning method has both 
metric-learning and meta-learning components,
which are learned in two stages.
The metric-learning stage learns 
both absolute and relative feature sets and then 
uses the Mahalanobis distance metric to compute class labels of the test sample.
The idea of using relative features stemmed from our prior work 
in domain adaption~\cite{das2018sample,das2018unsupervised,dasICANN}. 
Domain adaptation considers adaptation between labeled source-domain data 
and unlabeled target-domain data but with the same categories in both domains. 
The meta-learning stage learns auxiliary knowledge for classification,  
which is a transformation from a sample to its corresponding class prototype. 
This idea is related to the work of Wang et al.~\cite{wang2016M}, 
where they learned to transform small-sample-model parameters 
to large-sample-model parameters. 
The work on few-shot learning without forgetting~\cite{gidaris2018dynamic} 
also used a category-agnostic transformation 
but with a different distance metric and without any procedure 
to avoid negative transfer.
The overall framework of our proposed few-shot learning approach 
is shown in Fig.~\ref{fig:whole}.

Our proposed few-shot learning image recognition system  
is trained using a large database of  $N_{base}$ base categories, 
which consists of a large number of samples from each category. 
Each of these categories contains a large amount of data 
that we can use to learn some useful generalizable knowledge. 
This knowledge should help the recognition of $N_{novel}$ novel categories 
for which only a few labeled samples per category are available. 

The knowledge can be learned using traditional supervised learning, 
where training is generally carried out by feeding instances 
from the base categories in the form of mini-batches 
and then optimizing some loss function. 
The model is generally tested on the same set of categories 
on which it is trained. 
If we want the trained model to work on novel categories, 
then the model can be fine-tuned 
on the new training dataset~\cite{oquab2014learning}. 
However, the procedure of fine-tuning might not work 
if the novel categories have very few samples in each category. 
In fact, the fine-tuning procedure might cause the model to overfit 
on the few training samples, causing it to under-perform on novel category test samples. 
The main reason for overfitting is that the number of training samples per category 
is much less compared to the dimensionality of the feature space 
and therefore the variance of the few samples is inaccurate 
to capture the distribution of the class. 

We address these shortcomings of high dimensionality and variable variance 
by proposing the use of relative features, variance estimator 
and category-agnostic transformation.
Still, the traditional training procedure involving mini-batches 
from a large dataset would not be able to produce a satisfactory model 
since it does not simulate the test condition well. 
Each test category contains only a few samples 
and extracting mini-batches for training is impossible. 
Hence, an \emph{episodic} training strategy 
inspired from~\cite{vinyals2016matching} needs to be deployed.

In episodic learning, 
the set of few labeled samples available from each of the novel categories 
is known as the \emph{support set}. 
The set of unobserved testing samples of the novel categories 
is often called the \emph{query set}. 
If the support set were large, we could have just trained the model on the support set. 
However, since the support set is small, 
traditional training of a model would result in over-fitting 
and consequently the model would produce unsatisfactory performance 
on the testing data. 
However, the episodic training strategy can avert poor performance 
by simulating the test conditions. 
In each training episode, 
we first select $N$ classes randomly from among the $N_{base}$ base categories. 
From each of those selected $N$ classes, 
we randomly select $K$ and $Q$ disjoint samples from it. 
This sampling strategy is called the $N$-way $K$-shot sampling strategy. 
In general, $K$ is same as the number of support samples present per novel category. $Q$ is user-specified and is generally set in the range of 5 to 15 per category. Using this $N$-way $K$-shot sampling strategy, 
we form the training support set $\mathcal{S}=\{(\bx_i,y_i)\}_{i=1}^{n_s}$, 
where $n_s=K\! \times \!N$, 
and also the training query set $\mathcal{Q}=\{(\bx_j,y_j)\}_{j=1}^{n_q}$, 
where $n_q=Q\! \times \!N$. 
In the training episode, 
the support set is used to represent the class while the query set 
is used for the evaluation. 

\subsection{Relative-Feature-Space Representation}
The first step of our proposed few-shot learning framework 
requires feature extraction from the raw samples. 
This is done by feeding the support set samples $\bx_i$ 
from $\mathcal{S}$ and the query set samples $\bx_j$ from $\mathcal{Q}$ 
through the feature extraction module $\mathbf{f}_{\mathbf{\phi}}$ 
to produce the embeddings $\mathbf{f}_{\mathbf{\phi}}(\bx_i)$ and $\mathbf{f}_{\mathbf{\phi}}(\bx_j)$, respectively.
The dimensionality of this absolute feature map 
$\mathbf{f}_{\mathbf{\phi}}$ is very large 
compared to the total number of support and query samples. 
This sparsity in the number of samples compared to the dimension volume 
generally leads to over-fitting and poor generalization performance. 
To address this dimensionality problem, 
we propose the relative-feature-space representation,
which has a  dimensionality comparable to the total number of support and query samples in an episode.
The dimensionality of this relative feature space will therefore be much less than 
the original absolute feature space.
 
The relative feature of a sample in an episode is computed by 
calculating the squared pairwise Euclidean distance with itself and to all other samples in the episode. 
Hence, if there are $r=n_s+n_q$ samples in an episode, 
counting all $n_s$ support and $n_q$ query samples regardless of the categories, 
then the $d^{th}$ dimension of the relative feature $\mathbf{f}_{\mathbf{\rho}}$ 
of a sample $\bx_k$ is given as 
\begin{equation}
\label{eqn:relative}
[\mathbf{f}_{\mathbf{\rho}}(\bx_k)]_d=||\mathbf{f}_{\mathbf{\phi}}(\bx_k)-\mathbf{f}_{\phi}(\bx_d)||^2_2,
\end{equation} 
where $k,d \in \{1,2,...,r\}$ and $||\cdot||_2$ is the Euclidean norm. 
Note that $[\mathbf{f}_{\rho}(\bx_k)]_d=0$ for $k=d$. 
The dimensionality of this relative feature map is therefore $r$. 
Since this relative feature-space dimensionality is comparable to the number of samples 
and that these features contain important structural information about the data, 
we expect that the inclusion of this feature 
would increase few-shot testing performance. 

\begin{figure}[]
\centering
\includegraphics[width=8cm]{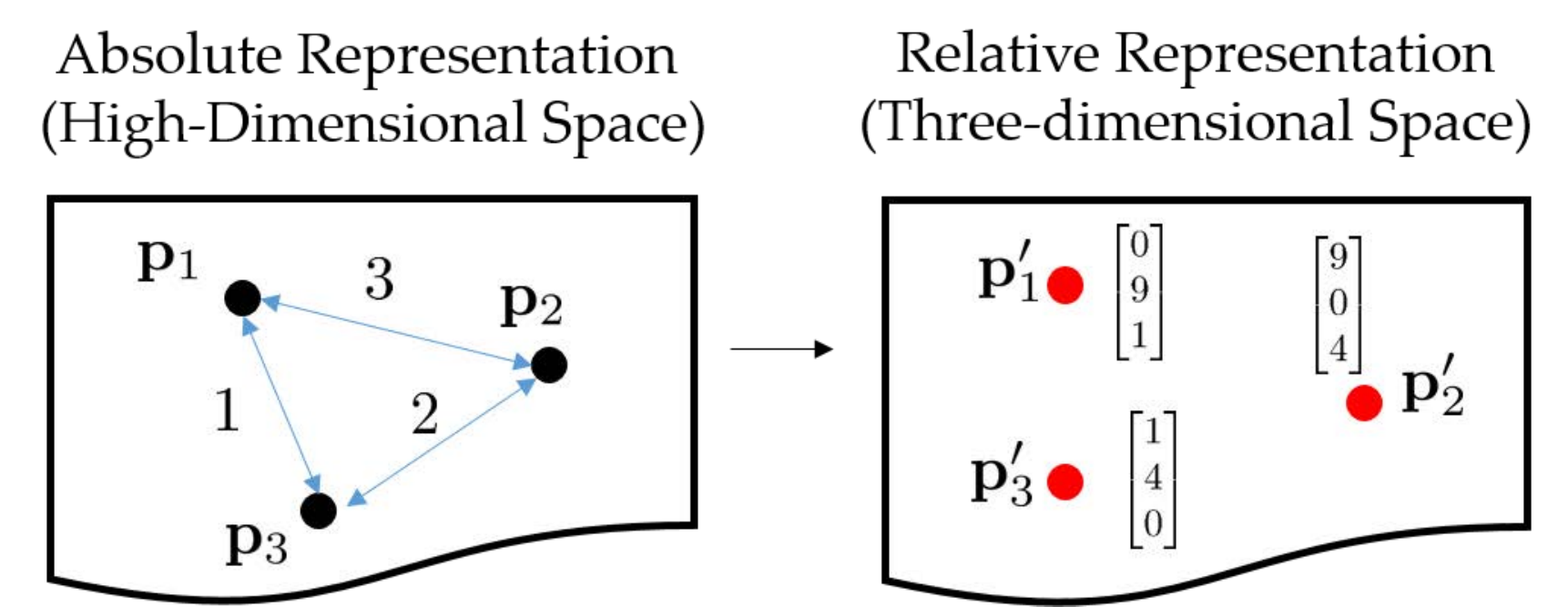}
\caption{This figure shows an example on how the low-dimensional relative-feature representation is computed from the original high-dimensional representation space. The original high dimensional feature space contains three data points. Accordingly, we would obtain a three-dimensional feature space if we compute pairwise distances of a data-point with itself and other points.}
\label{fig:relrep}
\end{figure}

In Fig.~\ref{fig:relrep}, we show a simple example 
on how to compute the relative-feature representation from the absolute-feature representation. 
Consider that there are three image samples -- $\bx_1$, $\bx_2$ and $\bx_3$ in an episode 
whose absolute-feature representations are $\bp_1=\mathbf{f}_{\phi}(\bx_1)$, 
$\bp_2=\mathbf{f}_{\phi}(\bx_2)$, and $\bp_3=\mathbf{f}_{\phi}(\bx_3)$, respectively. 
They are pairwise separated through Euclidean distances 
of 1, 2 and 3 as shown in the figure. 
From Eq.~\eqref{eqn:relative}, 
the relative-feature representation is obtained by squaring the pairwise Euclidean distances. 
Since there are three points in the episode, 
these points will lie in a three-dimensional relative-feature representation space 
and they would be represented as $\bp'_1=\mathbf{f}_{\rho}(\bx_1)=[0,9,1]^{T}$, $\bp'_2=\mathbf{f}_{\rho}(\bx_2)=[9,0,4]^{T}$ 
and $\bp'_3=\mathbf{f}_{\rho}(\bx_3)=[1,4,0]^{T}$.  

\subsection{Variance Estimation}
After embedding the support and query points 
in the absolute-feature space ($\mathbf{f}_{\phi}$) 
and the relative-feature space ($\mathbf{f}_{\rho}$), 
our goal is to use these features for classification. 
We do not want to tie our model to any category. 
We want to make our model generalizable to novel categories 
and therefore we do not use a classification layer 
that is commonly used for traditional neural networks. 
Instead, a nearest-class-mean approach is used~\cite{mensink2013distance}, 
where the query point embeddings are compared to 
the prototype representation of each class. 
The prototypes of a class $\bp_{r\phi_c}$ and $\bp_{r\rho_c}$ 
can be found by averaging the embedded support points 
of its class for both the absolute and relative representations, respectively, 
as follows
\begin{equation}
\label{eqn:protoabs}
\bp_{r\phi_c}=\frac{1}{|\mathcal{S}_c|}\sum_{(\bx_i,y_i) \in \mathcal{S}_c}\mathbf{f}_{\phi}(\bx_i) ,
\end{equation}   
\begin{equation}
\label{eqn:protorel}
\bp_{r\rho_c}=\frac{1}{|\mathcal{S}_c|}\sum_{(\bx_i,y_i) \in \mathcal{S}_c}\mathbf{f}_{\rho}(\bx_i),
\end{equation} 
where $\mathcal{S}_c$ is the set of samples of the support set $\mathcal{S}$, 
which belongs to class $c$. 
Using these prototypes, 
we can proceed to calculate the probability distribution 
over classes $p_{\phi}(y=c | \mathbf{x})$ and $p_{\rho}(y=c | \mathbf{x})$ 
for a query point $\mathbf{x}$. 
This is done using the softmax operation with distance metrics 
$d_{\phi}(\cdot)$ and $d_{\rho}(\cdot)$ 
for the absolute and relative representations, respectively, as follows
\begin{equation}
\label{eqn:probabs}
p_{\phi}(y=c | \mathbf{x})=\frac{\text{exp}(-d_{\phi}(\mathbf{f}_{\phi}(\mathbf{x}),\bp_{r\phi_c}))}{\sum_{c'}\text{exp}(-d_{\phi}(\mathbf{f}_{\phi}(\mathbf{x}),\bp_{r\phi_{c'}}))} ,
\end{equation}   
\begin{equation}
\label{eqn:probrel}
p_{\rho}(y=c | \mathbf{x})=\frac{\text{exp}(-d_{\rho}(\mathbf{f}_{\rho}(\mathbf{x}),\bp_{r\rho_c}))}{\sum_{c'}\text{exp}(-d_{\rho}(\mathbf{f}_{\rho}(\mathbf{x}),\bp_{r\rho_{c'}}))},
\end{equation} 
where the summation $\sum_{c'}$ is over all the classes present in the episode. 
In Eqs.~\eqref{eqn:probabs} and \eqref{eqn:probrel}, 
the distance metrics $d_{\phi}(\cdot)$ and $d_{\rho}(\cdot)$ 
need to be defined in order to compute the probability distributions. 
Snell et. al~\cite{snell2017prototypical} compared cosine and Euclidean distances 
and found Euclidean distance to perform better for few-shot testing. 
They argued that the Euclidean-distance metric 
is an example of Bregman Divergence.  
As a result, prototype computation and inference can be thought of 
as performing a mixture density estimation with exponential family distributions. 
However, if the Euclidean distance is used, 
we assume that all the classes have the same spread in the embedding space. 
This assumption may lead to poor classification performance 
because all the classes may not have the same variance.
Thus, we propose to use the Mahalanobis distance to measure 
and include the spread of each class in the classification scheme.

The Mahalanobis metric measures the distance 
between a data point $\mathbf{x}$ and a distribution $\mathcal{D}$. 
If the distribution $\mathcal{D}$ has an associated mean $\mu$ 
and an invertible covariance matrix $\mathbf{S}$, 
then the Mahalanobis distance $d_M$ is calculated as
\begin{equation}
\label{eqn:maha0}
d_M = \sqrt{(\mathbf{x}-\mu)^T\mathbf{S}^{-1}(\mathbf{x}-\mu)},
\end{equation} 
where  $\mathbf{S}^{-1}$ is the inverse of the covariance matrix $\mathbf{S}$.
In case the distribution is spherically Gaussian with a variance $\sigma^2$ 
for all the feature dimensions, 
the Mahalanobis distance $d_M$ is reduced to
\begin{align}
& d_M = \sqrt{(\mathbf{x}-\mu)^T\mathbf{S}^{-1}(\mathbf{x}-\mu)} = \sqrt{(\mathbf{x}-\mu)^T\mathbf{(\sigma^2\mathbf{I})}^{-1}(\mathbf{x}-\mu)} \nonumber \\ &=
\sqrt{\frac{(\mathbf{x}-\mu)^T(\mathbf{x}-\mu)}{\sigma^2}} = 
\sqrt{\frac{||\mathbf{x}-\mu||^2_2}{\sigma^2}}=
{\frac{||\mathbf{x}-\mu||_2}{\sigma}},
\label{eqn:maha1}
\end{align}
where $\mathbf{I}$ is an appropriate identity matrix.
 
\begin{figure}[]
\centering
\includegraphics[width=6cm]{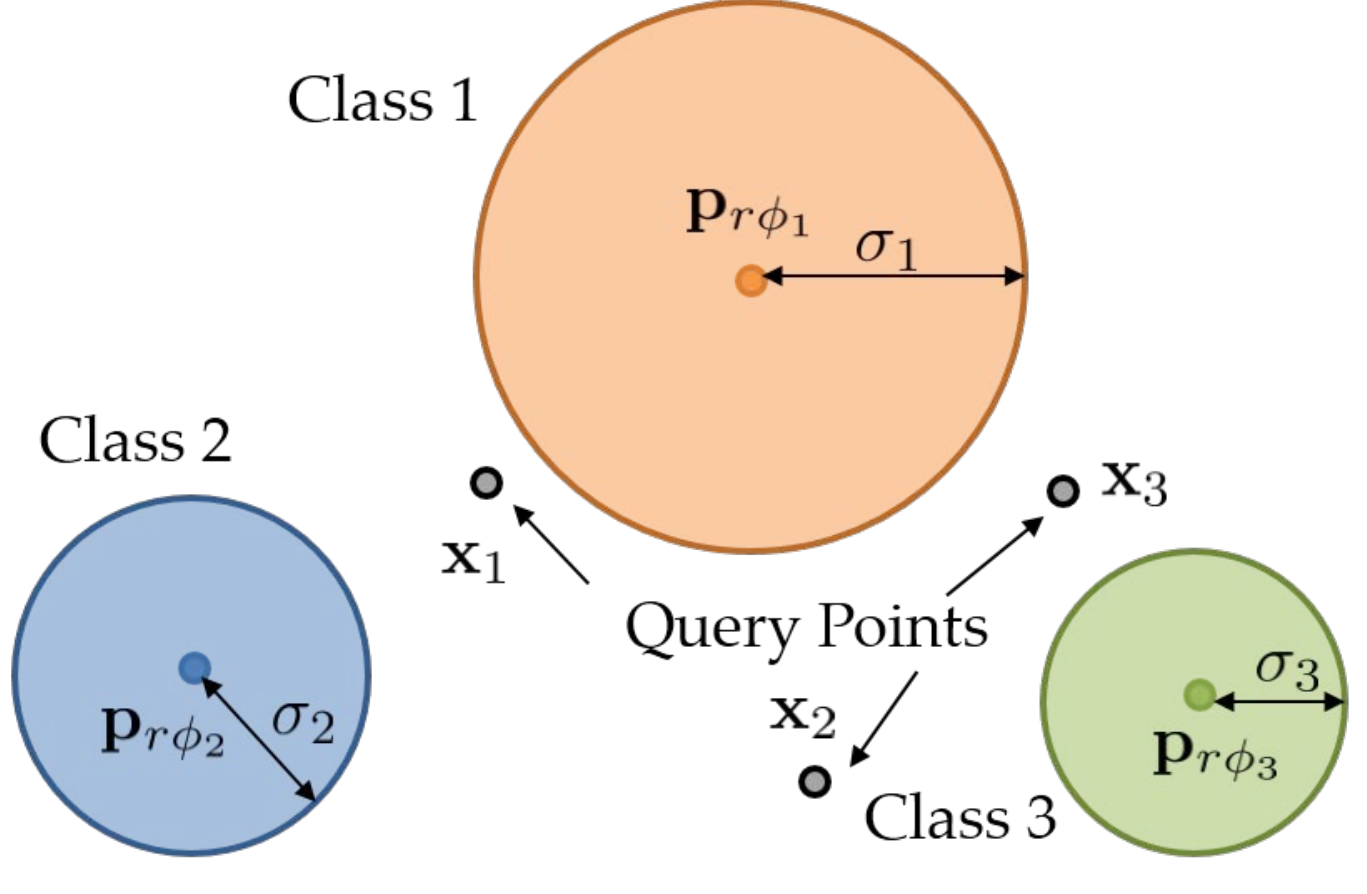}
\caption{This figure shows an example where different classes can have different variances. As a result, the Mahalanobis distance maybe preferred over Euclidean distance for classifying a test query point into one of these classes.}
\label{fig:proto}
\vspace*{-0.1in}
\end{figure}

The importance of using the Mahalanobis distance over the Euclidean distance 
is illustrated in Fig.~\ref{fig:proto} in which 
we have three classes with prototypes centered 
at $\mathbf{p}_{r\phi_1}$, $\mathbf{p}_{r\phi_2}$ and $\mathbf{p}_{r\phi_3}$. 
The spread of the classes is quantified through the standard deviations 
$\sigma_1$, $\sigma_2$ and $\sigma_3$.  
The goal is to classify the query points $\mathbf{x}_1$, $\mathbf{x}_2$ 
and $\mathbf{x}_3$ into one of the three classes. 
If we use the Euclidean distances for comparison, 
point $\mathbf{x}_1$ would yield equal probabilities for classes 1 and 2 
since the point is equidistant from those classes. 
This classification does not take into consideration 
that the spread of class 1 is more than the spread of class 2; 
that is, $\sigma_1> \sigma_2$. 
If we use the Mahalanobis distance, 
$\frac{||\mathbf{x}_1 - \mathbf{p}_{r\phi_1}||_2^2}{\sigma_1^2} < \frac{||\mathbf{x}_1 - \mathbf{p}_{r\phi_2}||_2^2}{\sigma_2^2} $, 
and accordingly the query point $\mathbf{x}_1$ 
will yield a higher probability for class 1. 
Similar treatment can also be applied to query points 
$\mathbf{x}_2$ and $\mathbf{x}_3$.

In our model, we expect each class to have its own covariance matrix $\mathbf{S}$. 
Therefore, there is a need to model the covariance $\mathbf{S}$ 
as a function of  each class's prototype. 
However, the covariance matrix $\mathbf{S} \in \mathbb{R}^{D\! \times \!D}$ 
is very high-dimensional, requiring lots of parameters to model it.  
Furthermore, the covariance matrix $\mathbf{S}$ is required to be positive definite, 
the constraints of which need to be satisfied strictly. 
Hence, we settle with using a spherical Gaussian distribution 
with the same variance for all the feature dimensions. 
Since we let the class variance be a function of the class's prototype, 
we can write
\begin{equation}
\label{eqn:variance}
\sigma_{c}^2=f_V(\bp_{r\phi_c}),
\end{equation}
where $\sigma_{c}^2$ and $\bp_{r\phi_c}$ 
are the variance and prototype of class $c$, respectively. 
This concept of predictable variance may be difficult to grasp initially. 
However, one can think of it as curve fitting of a function, 
where the input is the prototype and the output is the variance 
of the corresponding prototype. 
The corresponding function is fit using lots of data available 
from the base categories. 
Since we expect the function to be smooth, 
prototypes closer to each other should produce similar variances. 
After training is over, this function can then be used 
to predict the variance of novel-class prototypes.  
The variance estimating function $f_V$ can therefore be 
implemented by a neural network. 
Hence,  using Eqs.~\eqref{eqn:maha1} and \eqref{eqn:variance},
the distance metric $d_{\phi}(\cdot)$ in Eq.~\eqref{eqn:probabs} 
can be expressed as the square of the Mahalanobis distance as follows
\begin{equation}
\label{eqn:distabs}
d_{\phi}(\mathbf{f}_{\phi}(\mathbf{x}),\bp_{r\phi_c}) = \frac{||\mathbf{f}_{\phi}(\mathbf{x})-\bp_{r\phi_c}||^2_2}{\sigma_c^2}
= \frac{||\mathbf{f}_{\phi}(\mathbf{x})-\bp_{r\phi_c}||^2_2}{f_V(\bp_{r\phi_c})}.
\end{equation}
For the relative-feature space, 
the concept of having a variance does not have any physical meaning.  
As a result, we just use the square of the Euclidean distance metric for $d_{\rho}(\cdot)$ such that
\begin{equation}
\label{eqn:distrel}
d_{\rho}(\mathbf{f}_{\rho}(\mathbf{x}),\bp_{r\rho_c}) = ||\mathbf{f}_{\rho}(\mathbf{x})-\bp_{r\rho_c}||^2_2.
\end{equation}
The representation is learned by minimizing the negative log-probability 
averaged over all the query points. 
The negative log-probability of a query point is given as
\begin{equation}
L(\mathbf{\Phi},\bV) = - \log p_{\phi}(y=c | \mathbf{x}) -\lambda_{\rho}\log p_{\rho}(y=c | \mathbf{x}),
\label{eqn:reg}
\end{equation}
where $\mathbf{\Phi}$ and $\bV$ are composed of all the trainable parameters of the feature extractor ($\mathbf{f}_{\phi}$) and the variance estimator ($f_{V}$), respectively,
and $\lambda_{\rho}$ is a hyper-parameter 
for the regularization in Eq.~{\eqref{eqn:reg}}.
The negative log-probability averaged over all the query points 
in the batch needs to be minimized.

\subsection{Category-agnostic Transformation}
After the feature-extraction model and the variance estimator are trained, 
we proceed to the next stage of training. 
In this training stage, 
we propose to find a category-agnostic transformation 
from a mean-sample representation of a class 
to the prototype representation of the corresponding class. 
Learning this transformation is important 
because the novel categories have very few support samples 
and so the mean-sample representation will not accurately represent the prototype.
The existence of this category-agnostic transformation may be questionable. 
However, previous work by Wang et al.~\cite{wang2016M} 
suggested the existence of a similar transformation. 
In that work, the authors proposed the existence of a transformation 
between model parameters trained using less number of samples 
to model parameters trained using large number of samples. 
Since model parameters and samples are dual of each other, 
we conjecture the existence of a transformation 
between the mean-sample representation and the prototypes. 
We next determine this category-agnostic transformation and the factors 
that this transformation depends on. 
\begin{figure}[]
\centering
\includegraphics[width=6cm]{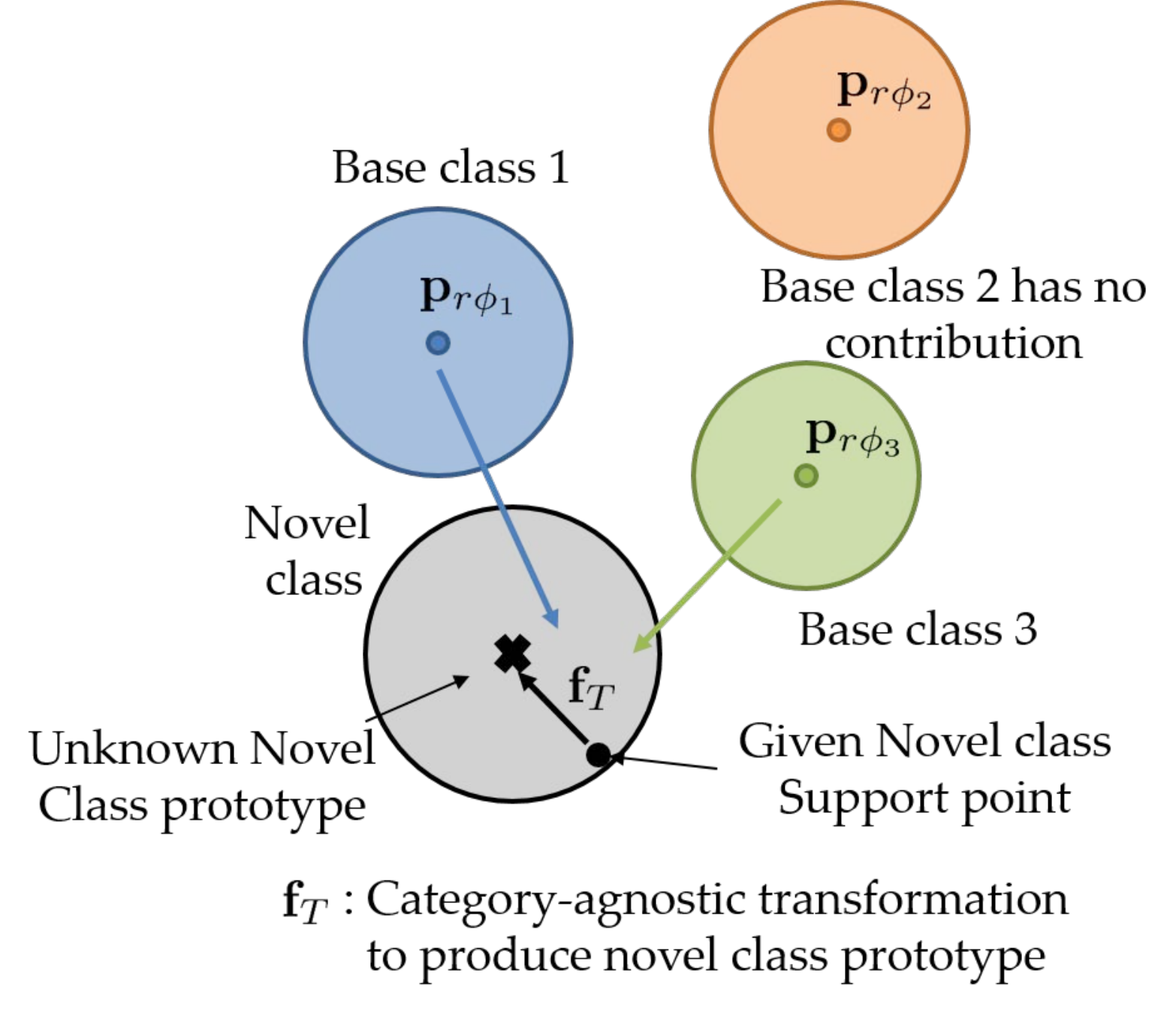}
\caption{Example depicting the choice of factors affecting the category-agnostic transformation from a support data-point to the corresponding prototype.}
\label{fig:category}
\vspace*{-0.1in}
\end{figure}

In addition to the mean-sample representation, 
the location of the novel-class prototype would also depend on the nearby base-class prototypes. 
This is illustrated through an example in Fig.~\ref{fig:category} 
in which we have one support sample point for a novel class. 
But this support data-point may not always be able to represent a class prototype 
because it might be present on the edge of the distribution as in this example. 
The transformation function mapping the support point 
to the unknown class prototype should depend on the support point
as well as on the nearby similar base categories. 
This is because the neighboring class prototypes 
condition the possible locations of the novel-class prototype. 
In this example, base classes 1 and 3 form the neighboring categories 
on which the location of the novel-class prototype should depend. 
Base class 2 is far from the novel class in the feature space 
and therefore it should have little effect on the location of the novel-class prototype.  
We next describe the construction of the transformation 
function $\mathbf{f}_\text{T}$. 

The prototype of a novel category $c$ depends on 
the mean-sample representation and the base-category prototypes 
collected in $\mathbf{P}_r$, 
where $\mathbf{P}_r \in \mathbb{R}^{n_p\! \times \!D}$ 
consists of the $n_p$ base-category prototypes stacked vertically in a matrix, 
and $D$ is the dimension of the absolute-feature space 
in which the prototypes lie. 
Ideally, the prototype matrix should be calculated using the base categories. 
Since each base category has a large number of samples, 
the mean representation will be used as an accurate estimate of the prototype. 
Thus, the prototype $\bp'_{r{\phi}_c}$ of a novel class $c$ can be represented as 
\begin{equation}
\label{eqn:prototrans}
\bp'_{r{\phi}_c}=\mathbf{f}_\text{T}(\bp_{r{
\phi_c}},\mathbf{P}_r),
\end{equation}
where $\bp_{r{\phi_c}}$ is the mean-sample representation of the novel class $c$. 
We can decompose the function $\mathbf{f}_\text{T}(\bp_{r\phi_c},\mathbf{P}_r)$
into two functions,
$\mathbf{f}_\text{T}(\bp_{r\phi_c},\mathbf{P}_r) =\mathbf{f}_{\text{T}_{1}}(\bp_{r\phi_c}) + \mathbf{f}_{\text{T}_{2}}(\bp_{r\phi_c},\mathbf{P}_r)$, 
where $\mathbf{f}_{\text{T}_{1}}$ is the contribution 
due to the mean-sample representation and 
$\mathbf{f}_{\text{T}_{2}}$ is the contribution 
due to the base-class prototypes $\mathbf{P}_r$. 
Since the contribution of the base-class prototypes 
depends on the closeness of $\bp_{r\phi_c}$ to the prototypes in 
$\mathbf{P}_r$, $\mathbf{f}_{\text{T}_{2}}$ will also depend on $\bp_{r\phi_c}$. 
We next discuss the construction of functions 
$\mathbf{f}_{\text{T}_{1}}$ and $\mathbf{f}_{\text{T}_{2}}$.

\textbf{Contribution of novel-class samples using residual connection.} 
The function $\mathbf{f}_{\text{T}_{1}}$ 
is a complex non-linear function 
that transforms the mean-sample representation $\bp_{r\phi_c}$ 
towards the prototype $\bp'_{r\phi_c}$. 
In case the number of samples in the novel category is large, 
$\bp_{r\phi_c}$ should identically map to $\bp'_{r\phi_c}$. 
Hence, it is important for the function $\mathbf{f}_{\text{T}_{1}}$ 
to model identity mappings. 
Residual connections and networks have been shown 
to model identity functions smoothly~\cite{he2016deep}. 
In our case, the corresponding meaningful residual connection will be 
$\mathbf{f}_{\text{T}_{1}}(\bp_{r\phi_c}) = \bp_{r\phi_c} + \mathbf{f}_{\text{T}_{11}}(\bp_{r\phi_c})$,
where  $\mathbf{f}_{\text{T}_{11}}(\bp_{r\phi_c})$ is a bias term 
and does not have a scaling effect on the mean-sample representation. 
Thus, if we include a scaled residual connection, then
\begin{equation}
\label{eqn:scaleresid}
\mathbf{f}_{\text{T}_{1}}(\bp_{r\phi_c}) = \bp_{r\phi_c}\mathbf{W}_1 + \mathbf{f}_{\text{T}_{11}}(\bp_{r\phi_c}),
\end{equation}
where $\mathbf{W}_1 \in \mathbb{R}^{D\! \times \!D}$ is the scaling matrix. 
Letting $\mathbf{f}_{\text{T}_{12}}(\bp_{r\phi_c})=\bp_{r\phi_c}\mathbf{W}_1$, 
the bias term $\mathbf{f}_{\text{T}_{11}}(\bp_{r\phi_c})$ 
will be a complex non-linear term and can be modeled 
using a multi-layer neural network.

\textbf{Contribution of the base classes.} 
The function $\mathbf{f}_{\text{T}_{2}}$ models 
the contribution of base-class prototypes to the novel-class prototype. 
Base classes that are similar to the novel class will have more contribution. 
This similarity can be measured in terms of Euclidean distance 
between a novel class mean-sample representation and a base-class prototype. 
The contribution of a base class $l$ to a novel class $c$ 
is quantified through a probability distribution,
\begin{equation}
\label{eqn:probe}
p_{p}(c,l)=\frac{\text{exp}(-||[\mathbf{P}_r]_l-\bp_{r\phi_c}||^2_2)}{\sum_{l'}\text{exp}(-||[\mathbf{P}_r]_{l'}-\bp_{r\phi_c}||^2_2)},
\end{equation}
where $[\mathbf{P}_r]_l$ is the prototype belonging to the $l^{th}$ base class. 
The computation of probability is carried out for all the base classes $l=1,2,...,n_p$. 
These are stacked together to form a probability vector $\mathbf{p}_c$ 
for the novel class $c$. 
After that, we use a threshold $t_h$ on the probability vector 
$\mathbf{p}_c \in \mathbb{R}^{1\! \times \!n_p}$. 
Only the elements above the threshold $t_h$ are kept 
while other elements are set to zero. 
This thresholding step is important as it ignores the effect of base classes 
that have very little contribution to the novel classes. 
From the feature-space perspective, novel classes that are distant 
from the base classes are ignored. 
This step is our attempt to prevent negative transfer~\cite{pan2010survey}, 
where irrelevant base classes contributing to learning novel-class recognition 
will reduce the recognition performance. 
The thresholded probability vector is set as $\mathbf{p}^{t_h}_c$. 
This is used to combine the base-class prototypes such that
\begin{equation}
\label{eqn:ft2}
\mathbf{f}_{\text{T}_{2}}(\bp_{r\phi_c},\mathbf{P}_r) = \mathbf{p}^{t_h}_c\mathbf{P}_r\mathbf{W}_2,
\end{equation}
where $\mathbf{W}_2 \in \mathbb{R}^{D\! \times \!D}$ is the scaling matrix. 
The factor $\mathbf{p}^{t_h}_c\mathbf{P}_r$ linearly combines the contributing base-class prototypes. 
The presence of $\mathbf{W}_2$ is important in scaling the effect of this term to the whole transformation function $\mathbf{f}_{\text{T}}$. 
Next, we discuss the procedure to learn this category-agnostic transformation $\mathbf{f}_{\text{T}}$, using the large labeled dataset available from the base categories.

\textbf{Training Strategy.} 
In the second stage of training, 
we follow the \emph{episodic} training strategy similar to the first stage. 
In each training episode, 
we randomly sample $N_{pn}$ categories from among the $N_{base}$ categories. 
We call these $N_{pn}$ categories as pseudo-novel categories. 
We refer to the remaining $N_{base}-N_{pn}$ categories as pseudo-base categories. 
The goal of this training strategy is to simulate the testing scenario 
where we have novel classes as well as already known base classes. 

In a training episode, the prototypes of the pseudo-base categories 
are calculated using the mean-sample representation. 
These prototypes can be stacked together to form 
the prototype matrix $\mathbf{P}_r$. 
For each pseudo-novel category, 
we randomly select $K_{pn}$ and $Q_{pn}$ disjoint samples. 
From this, we form the training support set 
$\mathcal{S}_{pn}=\{(\bx_i,y_i)\}_{i=1}^{m_{pn}}$, 
where $m_{pn}=K_{pn}\! \times \!N_{pn}$ and also the training query set 
$\mathcal{Q}_{pn}=\{(\bx_j,y_j)\}_{j=1}^{n_{pn}}$, 
where $n_{pn}=Q_{pn}\! \times \!N_{pn}$.  
For a category $c$ belonging to one of the $N_{pn}$ categories, 
we calculate the corresponding class prototype $\mathbf{p}'_{r\phi_{c}}$ 
using Eqs.~\eqref{eqn:prototrans}-\eqref{eqn:ft2}. 
Using this modified prototype $\mathbf{p}'_{r\phi_{c}}$, 
we can proceed to calculate the probability distribution over classes 
for a query point $\mathbf{x}$. 
This is done using the softmax operation 
with the Mahalanobis distance metric as described previously
\begin{equation}
\label{eqn:probabsmod}
p'_{\phi}(y=c | \mathbf{x})=\frac{\text{exp}(-d_{\phi}(\mathbf{f}_{\phi}(\mathbf{x}),\bp'_{r\phi_c}))}{\sum_{c'}\text{exp}(-d_{\phi}(\mathbf{f}_{\phi}(\mathbf{x}),\bp'_{r\phi_{c'}}))},
\end{equation}    
where the summation $\sum_{c'}$ is over all the $N_{pn}$ 
pseudo-novel classes present in the episode. 

The training is carried out by minimizing the negative log-probability 
averaged over all the query points. 
The negative log-probability of a query point is given as 
$L(\mathbf{\Theta}) = - \log p'_{\phi}(y=c | \mathbf{x})$, 
where $\mathbf{\Theta}$ consists of the scaling matrices $\mathbf{W}_1$, 
$\mathbf{W}_2$ and all the trainable parameters 
of the residual network $\mathbf{f}_{\text{T}_{11}}$. 
We also include a regression-based regularization 
involving the ground truth and predicted prototypes of 
these $N_{pn}$ pseudo-novel classes. 
If the ground truth prototype of class $c$ is $\bp_{r\phi_c}^{gt}$ 
and the predicted prototype is $\bp'_{r\phi_c}$, 
then the corresponding regularization becomes 
$L_r(\mathbf{\Theta})=||\bp'_{r\phi_c} - \bp_{r\phi_c}^{gt}||_2^2$. 
This regularization is averaged over all the prototypes of pseudo-novel classes. 
The regularization coefficient is set as $\lambda_r$. 

\begin{algorithm}[]
\SetAlgoLined
 \textbf{Given:} Base category training data $\mathcal{D}=\{(\bx_i,y_i)\}_{i=1}^{n}$ where each $y_i \in \{1,2,\dots,C\}$. $\mathcal{D}_c$ is a subset of $\mathcal{D}$ containing elements from class $c$
 \\
 \textbf{Parameters:} $\lambda_{\rho}, \lambda_r$\\
Randomly initialize parameters of feature extraction ($\mathbf{\Phi}$) and variance estimation ($\mathbf{V}$) \\
\textbf{for} each episode\\
\quad $\mathcal{N} \leftarrow \text{Sample}(\{1,2,\dots,C\},N)$\\
\quad \textbf{for} $c \in \{1,2,\dots,N\}$\\
\quad \quad $\mathcal{S}_c \leftarrow \text{Sample}(\mathcal{D}_{\mathcal{N}_{c}},K)$\\
\quad \quad $\mathcal{Q}_c \leftarrow \text{Sample}(\mathcal{D}_{\mathcal{N}_{c}}\setminus \mathcal{S}_c ,Q)$\\
\quad \quad $\bp_{r\phi_c}=\frac{1}{|\mathcal{S}_c|}\sum_{(\bx_i,y_i) \in \mathcal{S}_c}\mathbf{f}_{\phi}(\bx_i)$ \\
\quad \quad $\bp_{r\rho_c}=\frac{1}{|\mathcal{S}_c|}\sum_{(\bx_i,y_i) \in \mathcal{S}_c}\mathbf{f}_{\rho}(\bx_i)
$\\
\quad \quad $\sigma_{c}^2=f_V(\bp_{r\phi_c})$\\
\quad \textbf{end for}\\
\quad $L_1 \leftarrow 0$\\
\quad \textbf{for} $c \in \{1,2,\dots,N\}$\\
\quad \quad \textbf{for} $(\mathbf{x},y) \in \mathcal{Q}_c$\\
\quad \quad \quad $L_1 \leftarrow L_1+\frac{1}{NQ}[(d_{\phi}(\mathbf{f}_{\phi}(\mathbf{x}),\bp_{r\phi_c}))+$\\
\qquad \quad \quad \quad \quad $\log (\sum_{c'}\text{exp}(-d_{\phi}(\mathbf{f}_{\phi}(\mathbf{x}),\bp_{r\phi_{c'}}))))+$\\
\qquad \quad \quad \quad \quad $\lambda_{\rho}(d_{\rho}(\mathbf{f}_{\rho}(\mathbf{x}),\bp_{r\rho_c}))+$\\
\qquad \quad \quad \quad \quad $\log (\sum_{c'}\text{exp}(-d_{\rho}(\mathbf{f}_{\rho}(\mathbf{x}),\bp_{r\rho_{c'}}))))]$\\
\quad \quad \textbf{end for}\\
\quad \textbf{end for}\\
Take gradient step of $L_1$ with respect to $\mathbf{\Phi},\mathbf{V}$\\
\textbf{end for}\\
First training stage ends and second training stage starts.\\ 
Randomly initialize parameters of category-agnostic transformer ($\mathbf{\Theta}$)\\
\textbf{for} each episode \\
\quad $\mathcal{N}_{nov} \leftarrow \text{Sample}(\{1,2,\dots,C\},N_{pn})$\\
\quad $\mathcal{N}_{base} \leftarrow \{1,2,\dots,C\} \setminus \mathcal{N}_{nov}$\\
\quad $\mathbf{P}_r \leftarrow \mathcal{N}_{base}$ [Form pseudo-base prototypes]\\
\quad $\mathbf{P}_{rn} \leftarrow \mathcal{N}_{nov}$ [Form pseudo-novel prototypes]\\
\quad \textbf{for} $c \in \{1,2,\dots,N_{pn}\}$\\
\quad \quad $\mathcal{S}_{pn_c} \leftarrow \text{Sample}(\mathcal{D}_{\mathcal{N}_{nov_c}},K_{pn})$\\
\quad \quad $\mathcal{Q}_{pn_c} \leftarrow \text{Sample}(\mathcal{D}_{\mathcal{N}_{nov_c}}\setminus \mathcal{S}_{pn_c} ,Q_{pn})$\\
\quad \quad $\bp'_{r\phi_c}=\mathbf{f}_\text{T}(\bp_{r\phi_c},\mathbf{P}_r)$\\
\quad \textbf{end for}\\
\quad $L_2 \leftarrow 0$\\
\quad \textbf{for} $c \in \{1,2,\dots,N_{pn}\}$\\
\quad \quad \textbf{for} $(\mathbf{x},y) \in \mathcal{Q}_{pn_c}$\\
\quad \quad \quad $L_2 \leftarrow L_2+\frac{1}{N_{pn}Q_{pn}}[(d_{\phi}(\bbf_{\phi}(\mathbf{x}),\bp'_{r\phi_c}))+$\\
\qquad \quad \quad \quad \quad $\log (\sum_{c'}\text{exp}(-d_{\phi}(\bbf_{\phi}(\mathbf{x}),\bp'_{r\phi_{c'}}))))+$\\
\qquad \quad \quad \quad \quad $\lambda_{r}||\bp'_{r\phi_c} - \bp_{r\phi_c}^{gt}||_2^2]$, where $\bp_{r\phi_c}^{gt}=[\mathbf{P}_{rn}]_c$\\

\quad \quad \textbf{end for}\\
\quad \textbf{end for}\\
Take gradient step of $L_2$ with respect to $\mathbf{\Theta}$\\
\textbf{end for}\\
\caption{Proposed two-stage few-shot learning procedure.}
\end{algorithm}

After the training is done, 
testing is also carried out in an episodic fashion. 
For each episode, 
we randomly sample $N_{test}$ classes from the novel test classes. 
From each novel class, 
$K_{test}$ support samples and $Q_{test}$ query samples are drawn randomly. 
The class prediction for a query point $\mathbf{x}$ 
is given as the class $c$ which minimizes $-\log p'_{\phi}(y=c | \mathbf{x}) -\lambda_{\rho}\log p_{\rho}(y=c | \mathbf{x})$.
The overall training procedure of the proposed two-stage 
few-shot learning method is provided in Algorithm 1. 

\begin{figure}[b]
\centering
\includegraphics[width=9cm]{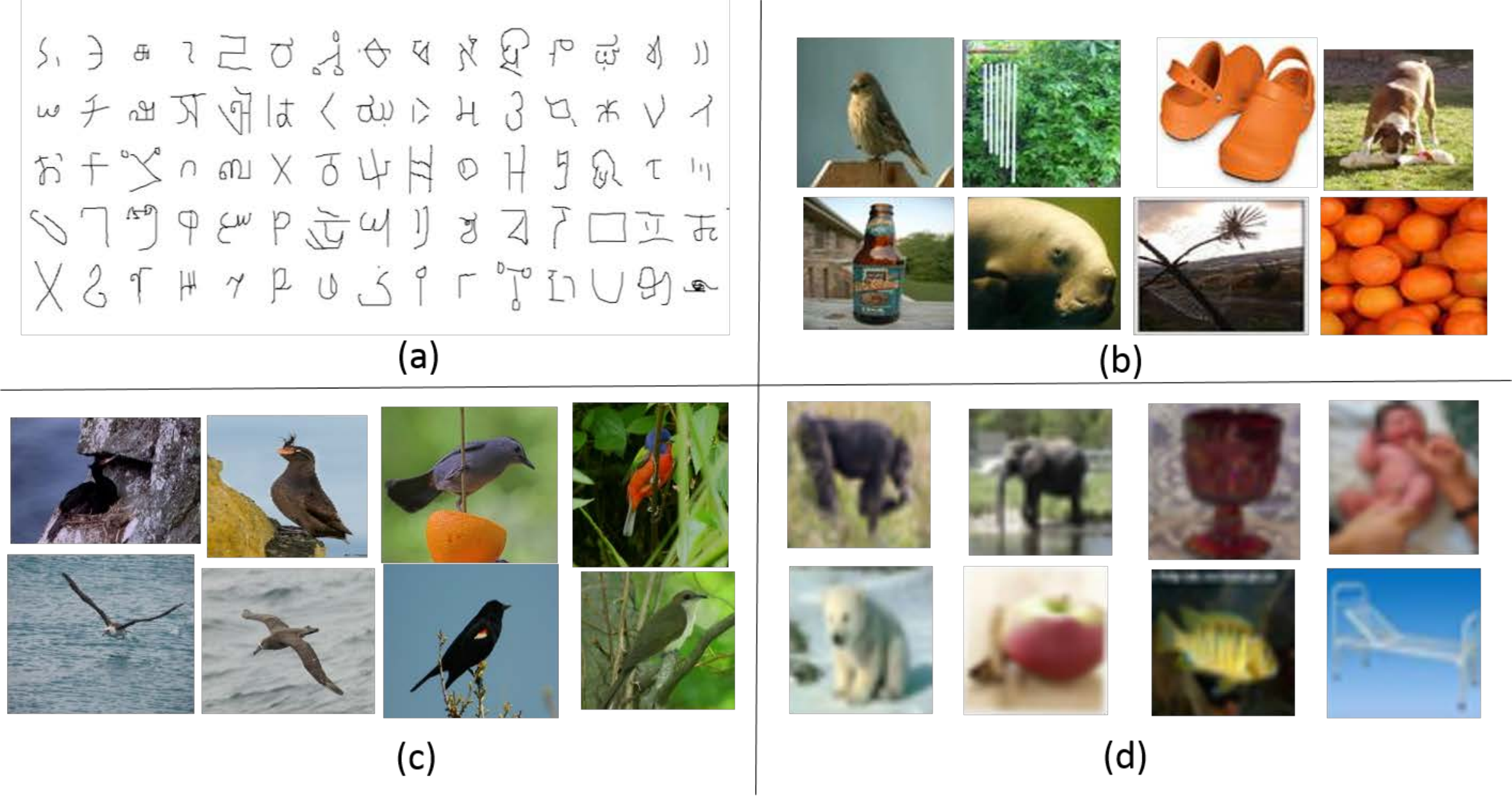}
\vspace*{-0.1in}
\caption{Instances of the dataset used in our experiment for 
(a) Omniglot, (b) miniImagenet, (c) CUB-200, and (d) CIFAR-100.}
\label{fig:dataset}
\vspace*{-0.1in}
\end{figure}

\section{Experimental Results}
\subsection{Datasets}
To evaluate our proposed few-shot learning approach, 
we performed experiments on four datasets -- the Omniglot~\cite{lake2011one}, 
the miniImagenet, the CUB-200~\cite{welinder2010caltech} 
and the CIFAR-100~\cite{cifar100} datasets. These datasets provide a large variety of category-level granularity, image resolution and categories to test upon. 
The Omniglot dataset consists of 1623 handwritten characters taken 
from 50 alphabets. 
Each character has 20 examples associated with it. 
Each example is written by a different person, 
resulting in sufficient intra-class variation. 
According to the procedure of Vinyals et al.~\cite{vinyals2016matching}, 
the images are resized to $28\!\times\!28$. 
Each character class is augmented with more samples 
by having rotations in multiples of 90 degrees. 
So around 1200 character classes (total of 4800 including rotations) 
are chosen as the training (i.e., base) categories 
and the remaining classes are chosen as the testing (i.e., novel) categories. 
The miniImagenet dataset is a subset of 
the ILSVRC-12 dataset~\cite{russakovsky2015imagenet}. 
It consists of RGB color images of size $84\! \times \!84$, 
consisting of 100 classes with 600 examples in each class. 
The 100 classes are divided into 64 for training (base), 
16 for validation and 20 for testing (novel). 

\begin{figure*}[t]
\centering
\includegraphics[width=15cm]{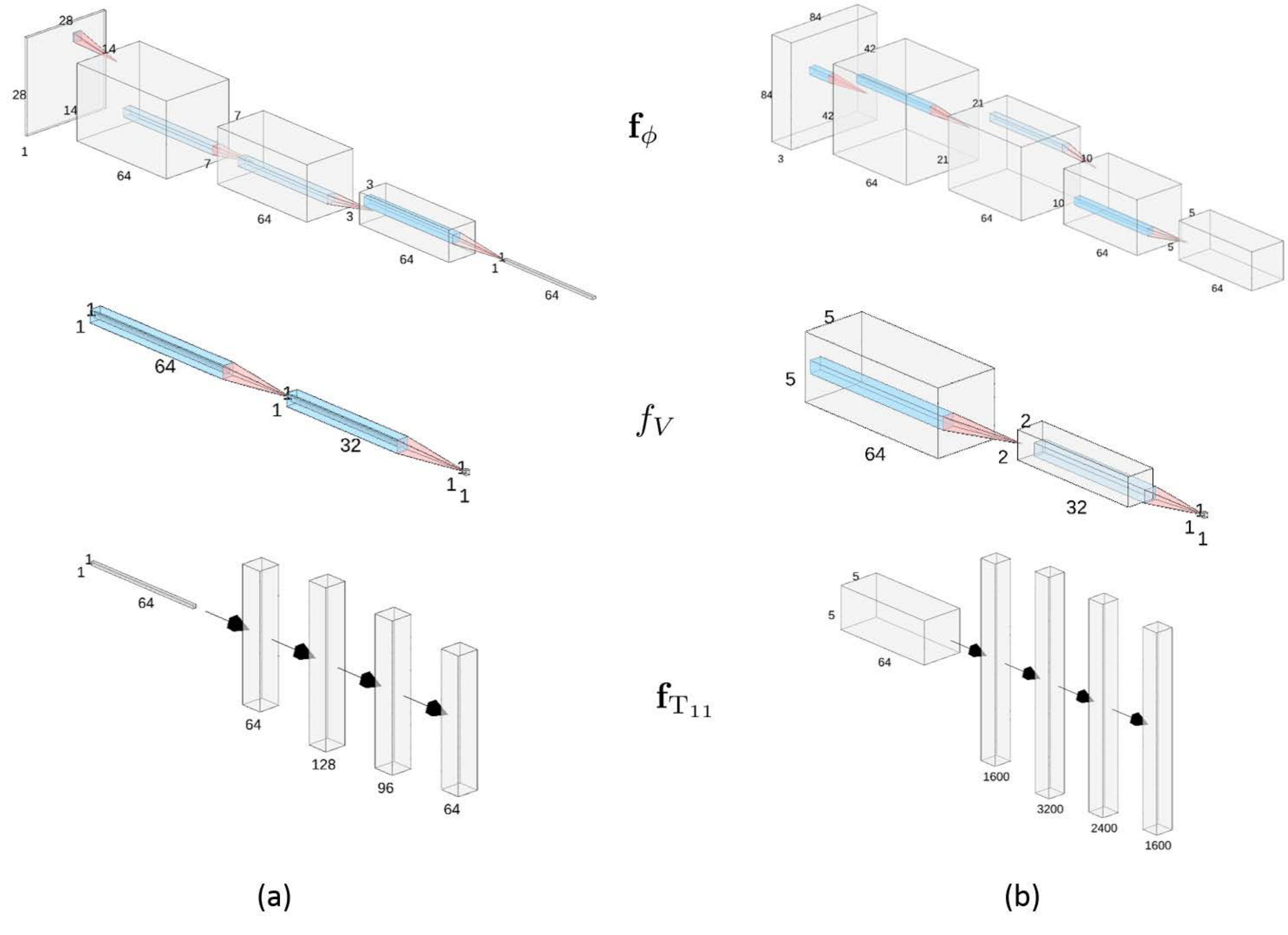}
\caption{Network architecture used for different modules $\bbf_{\phi}$, $f_V$ and $\bbf_{\text{T}_{11}}$. 
(a) For the Omniglot dataset,  
$\bbf_{\phi}$ produces a  $1\!\times\!64$ dimensional feature map 
from a $28\! \times \! 28 \! \times \! 1$ dimensional input image. 
The $f_V$ module produces a scalar variance from the feature map. 
The $\bbf_{\text{T}_{11}}$ regresses a 64-dimensional output from the feature map.
(b) For the miniImagenet dataset, 
$\bbf_{\phi}$ produces a $5\! \times\! 5 \!\times\! 64$ dimensional feature map 
from a $84\! \times\! 84 \!\times\! 3$ dimensional input image. 
The $f_V$ module produces a scalar variance from the feature map. 
The $\bbf_{\text{T}_{11}}$ regresses a 1600-dimensional output 
from the feature map.}
\label{fig:network}
\end{figure*}

The CUB-200 and CIFAR-100 datasets 
have been introduced long before but have only recently been used 
as a benchmark for few-shot learning algorithms. 
The CUB-200 dataset is a fine-grained dataset consisting of 11,788 images 
of size $84\! \times \!84\! \times \!3$, 
distributed across 200 categories of bird species. 
Using the class splits in~\cite{maco}, 
we have 100, 50 and 50 categories used for training, 
validation and testing, respectively. 
The CIFAR-100 dataset consists of 60000 low-resolution images 
of size $32\! \times \!32\! \times \!3$. 
These images are distributed across 100 fine-grained categories 
or 20 coarse-grained categories.  
Using the class splits in~\cite{bertinetto2016learning}, 
we have 64, 16 and 20 categories used for training, 
validation and testing, respectively. Figures~\ref{fig:dataset}(a), (b), (c) and (d) show some of the examples from the Omniglot, the miniImagenet, the CUB-200 and the CIFAR-100 datasets, respectively.

\subsection{Implementation}
In this sub-section, we discuss the details of our neural network architecture 
and the training procedure. 
For the feature extractor module ($\mathbf{f}_{\phi}$)
of our trainable neural network architecture, we use four convolutional blocks. 
{This feature extractor architecture is the same 
as used in previous works~\cite{snell2017prototypical,vinyals2016matching}. 
This is done for the sake of fair comparison. 
Most of these previous works selected the feature-extraction architecture empirically. 
For shallow convolutional architecture and therefore more high-dimensional feature space, the performance is poor because the features extracted are not robust and not class-discriminative enough.  But, as the depth of the convolutional architectures increases to a certain limit, we obtain a more informative low-dimensional feature space and therefore better recognition performance. 
The authors of~\cite{snell2017prototypical,vinyals2016matching} experimented 
and found that the presented four-convolutional-blocks-based 
architecture is lightweight and optimal.}
Each of these blocks consists of a 64-filter $3\!\times\! 3$ 
convolution layer with SAME padding, 
batch normalization layer, and an ReLU activation followed by 
a $2\!\times\! 2$ max-pooling layer all stacked upon another. 
The batch normalization~\cite{ioffe2015batch} results 
in better recognition performance because it prevents internal covariate shift. 
When a $28\! \times \!28$ Omniglot image is applied 
as an input to these four convolutional blocks, 
its output results in a $64$-dimensional feature space. 

The variance estimator $f_{V}$ consists of two convolutional blocks. 
Each convolutional block consists of 
$1\! \times \!1$ convolution layer with SAME padding, 
batch normalization layer and an ReLU activation layer. 
The first and the second convolutional blocks consist of 32 and 1 filters, respectively. 
The last layer producing the variance has softplus operation 
as the activation function. 
This is selected to produce only positive outputs. 

The transformation layer $\mathbf{f}_{\text{T}_{11}}$ 
consists of three fully connected layers of 128, 96 and 64 dimensions. 
Except the last layer, all the layers contain batch-normalization 
and ReLU activation functions. 
The last layer does not have an ReLU activation 
so that it can provide both negative and positive transformation shifts as output. 
The overall architecture of all the modules used for the Omniglot dataset 
is shown in Fig.~\ref{fig:network}(a). 

The neural-network structure was trained 
using the stochastic gradient descent variant \emph{Adam}~\cite{kingma2014adam} 
with an initial learning rate of $10^{-3}$. 
The first-stage training was carried out using 60-way 5-shot 
with 5 query points per episode. 
The higher way is chosen in training so that the model can learn 
a more difficult task of distinguishing more classes 
and therefore produce a more discriminative feature space. 
In this paper, the second-stage training episodic setup 
is always kept the same as the testing episodic setup for all the experiments; 
that is, if the testing setup is $N$-way $K$-shot, so is the second-stage training setup. 

The hyper-parameters $\lambda_\rho$, $\lambda_r$, and $t_h$ were set 
to $0.1$, $10^{-4}$, and $0.02$, respectively. 
It is important to note that these hyper-parameters are kept fixed 
for a particular dataset. 
This is mainly because cross-validation is not always feasible 
for the few-shot learning setting, 
which contains only a few samples from the target category. 
Also, the validation classes are not representative of the test classes.

For reporting the recognition performance, 
1000 random test episodes were selected and accuracy 
was obtained by averaging over all the test episodes.  
Each episode contained the corresponding $N$-way $K$-shot support samples 
and 5 query samples per way for testing.

\begin{table*}[t]
\caption{Results of few-shot classification on the Omniglot dataset. Accuracies in \% are reported as averaged over 1000 test episodes. Some of the studies report 95\% confidence interval while some do not report results as shown by '-' }
\label{tab:omni}
\centering
\begin{tabular}{@{}ccccc@{}}
\toprule
\textbf{Method}     & 5-way 1-shot & 5-way 5-shot & 20-way 1-shot & 20-way 5-shot \\ \midrule
SIAMESE~\cite{koch2015siamese}             & 97.3         & 98.4         & 88.1          & 97.0          \\
MANN~\cite{santoro2016meta}               & 82.8         & 94.9         & -             & -             \\
MATCHING NET~\cite{vinyals2016matching}       & 98.1         & 98.9         & 93.8          & 98.5          \\
SIAMESE MEMORY~\cite{rare}      & 98.4         & 99.6         & 95.0          & 98.6          \\
NEURAL STATISTICIAN~\cite{edwards2016towards} & 98.1         & 99.5         & 93.2          & 98.1          \\
MAML~\cite{finn2017model}               & 98.7$\pm$0.4  & \textbf{99.9}$\pm$\textbf{0.1}     & 95.8$\pm$0.3      & 98.9$\pm$0.2      \\
META NET~\cite{munkhdalai2017meta}           & 99.0         & -            & 97.0          & -             \\
PROTO NET~\cite{snell2017prototypical}           & 98.8         & 99.7         & 96.0          & 98.9          \\
RELATION NET~\cite{sung2018learning}        & \textbf{99.6}$\pm$\textbf{0.2}   & 99.8$\pm$0.1     & \textbf{97.6}$\pm$\textbf{0.2}     & \textbf{99.1}$\pm$\textbf{0.1}      \\
OUR PROPOSED METHOD                & 99.2$\pm$0.3 & 99.5$\pm$0.2  & 97.2$\pm$0.3         & 98.9$\pm$0.3  \\  
\bottomrule  
\end{tabular}
\end{table*}

For the miniImagenet dataset, 
we used the same feature extraction network architecture as the Omniglot dataset. 
However, since the miniImagenet dataset 
has images of size $84\! \times \! 84 \!\times \!3$, 
the convolution module produces a 1600-dimensional feature vector. 
The variance estimator is also the same as that of the Omniglot dataset 
except that this estimator contains 
a $2\!\times\!2$ max-pooling stage before the non-linearity. 
This is required to reduce the $5\!\times\!5\times\!64$ 
(1600-dimensional) feature map to a scalar variance value. 
The transformation layer $\mathbf{f}_{\text{T}_{11}}$ 
consists of three fully connected layers of 3200, 2400 and 1600 dimensions. 
The overall architecture of all the modules used for the miniImagenet dataset 
is shown in Fig.~\ref{fig:network}(b). 

The hyper-parameters $\lambda_\rho$, $\lambda_r$ and $t_h$ 
were set to $0.1$, $10^{-4}$ and $0.02$, respectively.  
For testing on the 5-way 1-shot and 5-way 5-shot episodic strategy, 
we used a 20-way 1-shot and 20-way 5-shot sampling strategy, respectively, 
in the first-stage training. 
Each episode contained the corresponding $N$-way $K$-shot 
support samples and 15 query samples per way for testing. 
Results were reported by computing the average accuracy 
over 600 such randomly sampled episodes with 95\% confidence interval.

For the CUB-200 and CIFAR-100 datasets, 
we used the same {four}-convolutional-blocks-based architecture 
as the feature extractor that has been previously used 
on the miniImagenet and Omniglot datasets. 
This embedding results in 1600 and 256 dimensional feature spaces 
for the CUB-200 and the CIFAR-100 datasets, respectively. 
The transformation layer $\mathbf{f}_{\text{T}_{11}}$ for the CUB-200 dataset
consists of three fully connected layers of 3200, 2400 and 1600 dimensions. 
The architecture of $f_{V}$ for the CUB-200 dataset is the same as that of the miniImagenet dataset.
The transformation layer $\mathbf{f}_{\text{T}_{11}}$ 
for the CIFAR-100 dataset
consists of three fully connected layers of 512, 384 and 256 dimensions. 
The architecture of $f_{V}$ for the CIFAR-100 dataset is similar to that of the miniImagenet dataset 
except that the $2\!\times\!2$ max-pooling step is applied only on the second convolutional block.   
The hyper-parameters $\lambda_\rho$, $\lambda_r$ and $t_h$ 
were set to $1.0$, $10^{-3}$ and $0.5$ 
on both the CUB-200 and the CIFAR-100 datasets. 
It is important to note that for a fair comparison, 
we only report previous work that used the simple four-convolutional-block-based 
embedding instead of the more sophisticated ResNet~\cite{he2016deep} architecture. 

\subsection{Comparison against Related Approaches}
Since our proposed few-shot learning method 
has both meta-learning and metric-learning components, 
we compared our proposed method against recent 
meta-learning~\cite{finn2017model,munkhdalai2017meta,ravi2016optimization} 
and metric-learning~\cite{koch2015siamese,vinyals2016matching,snell2017prototypical,sung2018learning} methods. 
We also compared against recent memory-based models~\cite{santoro2016meta,rare} 
and the Neural Statistician method~\cite{edwards2016towards} 
that learns how to represent statistics of the data.  
The results of the comparisons on the Omniglot dataset 
are shown in Table~\ref{tab:omni}.

As seen from Table~\ref{tab:omni}, 
most of the recent methods achieved almost perfect recognition performance 
on the Omniglot dataset 
(8 out of 10 methods obtained an average accuracy of 
more than 98\% for the 5-way 1-shot task). 
Our proposed method 
obtained an average accuracy of 99.2\% and 97.2\% 
for the 5-way 1-shot and 20-way 1-shot tasks, respectively, 
which are better than most of the previous approaches. 
However, Relational Network~\cite{sung2018learning} produced the best result; 
that is, 99.6\% and 97.6\% for the 5-way 1-shot and 20-way 1-shot tasks, 
respectively, because it learned a distance metric while our proposed method 
used a predefined Mahalanobis distance metric. 
The confidence interval of our proposed method (98.9\%-99.5\%) 
also overlapped with that of the Relational Network approach (99.4\%-99.8\%) 
for the 5-way 1-shot task. 
The confidence interval overlapped for the 20-way 1-shot task as well. 
As expected, higher shots during the testing 
produced better results (98.9\%$>$97.2\% for the 20-way task) 
for our proposed method because they represented 
the class statistics better than by just using one shot. 
Also, higher ways produced worse result (97.2\%$<$99.2\% for the 1-shot task) 
because there were more potential classes to choose from 
and the chances of misclassification were higher.

For the miniImagenet dataset, 
the comparison is more challenging and there is more room for improvement towards perfect performance. 
The results of the comparison are shown in Table \ref{tab:mini}. 
From Table \ref{tab:mini}, 
we can see that our proposed method produced an average accuracy 
of 52.68\% and 70.91\% on the 5-way 1-shot and 5-way 5-shot tasks, respectively, 
which are better than most of the previous methods. 
This can be mainly attributed to our two-stage training procedure, 
where the model learns to both represent and classify in a low-shot regime. {However, the methods -- {Predicting Parameters from Activation~\cite{qiao2018few} (PPA) and Transductive Propagation Networks~\cite{liu2018learning} (TPN)} produced better results 
than our proposed method in the 1-shot setting. 
Upon inspection, we realized that the PPA method used pre-trained embedding 
while most other {few-shot learning} methods and our training method 
of the embedding/feature extractor were done from scratch. 
Using a pre-trained embedding implies that datasets beyond the base 
and novel categories have been used in training the model 
and therefore the model would not be suitable for comparison.
However, we still included the results for PPA 
in Table \ref{tab:mini} for the sake of completeness. 
Also, the TPN method uses a transductive approach 
which assumes all the test/query data are available as a batch. 
The improvement in performance of this method is mainly 
due to the fact that the authors used the manifold of the unlabeled test data 
as well as support data to do inference. 
However, the method might not work if the number of query points is less 
or the query points arrive in a streaming fashion as in a real-world situation.}

{The results of our {proposed method} in comparison 
with previous work for the CUB-200 and CIFAR-100 datasets 
are shown in Table~\ref{tab:cub} and Table~\ref{tab:cifar}, respectively. 
In Table~\ref{tab:cub}, on the CUB-200 dataset, 
our proposed {method} produced {about} 6 points improvement 
over the second best method. 
Similarly, in Table~\ref{tab:cifar}, on the CIFAR-100 dataset,  
our proposed method produced around 2 points improvement over the second best method. 
This suggests that our proposed method can provide competitive performance 
on {fine-grained and low-resolution datasets} as well.  
Also, the average performance on the CUB-200 dataset 
is less than that on the CIFAR-100 dataset. 
This is because the CUB-200 dataset contains more fine-grained categories 
compared to the CIFAR-100 dataset and therefore classes overlap 
more in the CUB-200 dataset.} 

From these comparative studies, it is not clear how all the modules 
in our trainable neural-network architecture contributed to the performance. 
Therefore, we resort to further analyzing each component 
of our proposed {method} in the following sub-sections.

\begin{table}[th]
\caption{Results of few-shot classification on the miniImagenet dataset. Accuracies are reported as averaged over 600 test episodes. Most of these studies report 95\% confidence interval while unreported results   are shown as '--'} 
\label{tab:mini}
\centering
\begin{tabular}{@{}ccc@{}}
\toprule
\textbf{Method} & 5-way 1-shot   & 5-way 5-shot  \\ \midrule
META-LSTM~\cite{ravi2016optimization}       & 43.44$\pm$0.77     & 60.60$\pm$0.71 \\
MAML~\cite{finn2017model}            & 48.70$\pm$1.84     & 63.11$\pm$0.92    \\
MATCHING NET~\cite{vinyals2016matching}   & 43.56$\pm$0.84     & 55.31$\pm$0.73    \\
META NET~\cite{munkhdalai2017meta}        & 49.21$\pm$0.96     & --             \\
PROTO NET~\cite{snell2017prototypical}       & 49.42$\pm$0.78     & 68.20$\pm$0.66    \\
RELATION NET~\cite{sung2018learning}    & 51.38$\pm$0.82     & 67.07$\pm$0.69    \\
{GNN}~\cite{garcia2017few}    & 50.33$\pm$0.36 & 66.41$\pm$0.63  \\
{REPTILE}~\cite{reptile}   & 49.97     & 65.99    \\
{TPN}~\cite{liu2018learning}   & 53.75     & 69.43    \\
{PPA}~\cite{qiao2018few}    & \textbf{54.53$\pm$0.40}     & 67.07$\pm$0.20    \\
{R2D2}~\cite{bertinetto2018meta}   & 51.8$\pm$0.2     & 68.4$\pm$0.2    \\
OUR PROPOSED METHOD           & 52.68$\pm$0.51 & \bf{70.91$\pm$0.85} \\ \bottomrule
\end{tabular}
\end{table}

\begin{table}[th]
\caption{{Results of few-shot classification on the CUB-200 dataset where our accuracy is reported as averaged over 600 test episodes}} 
\label{tab:cub}
\centering
\begin{tabular}{@{}ccc@{}}
\toprule
\textbf{Method} & 5-way 1-shot   & 5-way 5-shot  \\ \midrule
META-LSTM~\cite{ravi2016optimization}       & 40.43     & 49.65 \\
MAML~\cite{finn2017model}            & 38.43     & 59.15    \\
MATCHING NET~\cite{vinyals2016matching}   & 49.34 & 59.31    \\
PROTO NET~\cite{snell2017prototypical}       & 45.27     & 56.35    \\
OUR PROPOSED METHOD           & \bf{55.85} & \bf{66.73} \\ \bottomrule
\end{tabular}
\end{table}

\begin{table}[th]
\caption{{Results of few-shot classification on the CIFAR-100 dataset where the accuracy is reported as averaged over 10000 test episodes. Most of these studies report 95\% confidence interval}} 
\label{tab:cifar}
\centering
\begin{tabular}{@{}ccc@{}}
\toprule
\textbf{Method} & 5-way 1-shot   & 5-way 5-shot  \\ \midrule
MAML~\cite{finn2017model}            & 58.9$\pm$1.9     & 71.5$\pm$1.0   \\
PROTO NET~\cite{snell2017prototypical}       & 55.5$\pm$0.7     & 72.0$\pm$0.6    \\
RELATION NET~\cite{sung2018learning}    & 55.0$\pm$1.0 & 69.3$\pm$0.8    \\
{GNN}~\cite{garcia2017few}    & 61.9 & 75.3  \\
{R2D2}~\cite{bertinetto2018meta}   & 65.4$\pm$0.2     & 79.4$\pm$0.2    \\
OUR PROPOSED METHOD           & \bf{67.15$\pm$0.3} & \bf{81.65$\pm$0.3} \\ \bottomrule
\end{tabular}
\end{table}

\begin{table*}[]
\caption{Ablative study of our approach on the miniImagenet dataset. 
Averaged accuracy is reported as the training way is varied. 
Ablations include the Variance estimator (V), 
Relative features (R), and Category-agnostic Transformer (T). 
The baseline is the Prototypical Network (PN)}
\label{tab:ablate}
\centering
\begin{tabular}{@{}lcccccc|cccccc@{}}
\toprule
          & \multicolumn{6}{c|}{5-way 1-shot Testing}                    &  \multicolumn{6}{c}{5-way 5-shot Testing}                    \\ \midrule
Training way & 5      & 10     & 15     & 20     & 25     & 30     & 5      & 10     & 15     & 20     & 25     & 30     \\ \midrule
PN        & 43.987 & 46.956 & 46.589 & 46.122 & 47.253 & 47.3   & 62.693 & 64.742 & 64.524 & 63.578 & 62.416 & 61.9   \\
PN+V      & 44.411 & 47.067 & 47.936 & 48.304 & 47.778 & 48.067 & 64.813 & 65.033 & 66.158 & 65.37  & 64.318 & 64.82  \\
PN+R      & 47.849 & 50.309 & 52.631 & 52.607 & 52.14  & 51.996 & 66.758 & 70.831 & 70.771 & 70.447 & 71.147 & 62.733 \\
PN+T      & 43.942 & 45.944 & 47.263 & 48.022 & 48.122 & 48.011 & 62.396 & 63.316 & 64.342 & 63.024 & 63.531 & 64.86  \\
PN+V+R    & 49.322 & 51.057 & 51.031 & 52.782 & 52.716 & 51.773 & 69.1   & 70.936 & 71.496 & 71.36  & 70.36  & 68.23  \\
PN+V+T    & 45.689 & 47.927 & 48.422 & 48.002 & 47.693 & 47.947 & 61.667 & 63.484 & 63.736 & 62.431 & 61.978 & 63.48  \\
PN+R+T    & 46.913 & 51.224 & 52.338 & 53.036 & 53.789 & 53.66  & 68.76  & 71.38  & 72.34  & 72.151 & 72.584 & 67.34  \\ \bottomrule
\end{tabular}
\end{table*}

\subsection{Ablation Study with Varying Training and Testing Conditions}
The contribution of this paper consists of the following modules 
on top of the Prototypical Network (PN) -- a variance estimator (V), 
the relative features (R), and the category-agnostic transformer (T). 
We thus performed an ablative study, 
where we added all combinations of the modules on the PN 
and observed the change in performance. 
Results of this experiment are reported in Table \ref{tab:ablate} 
as the training way is varied for the 5-way 1-shot 
and 5-way 5-shot testing conditions.

We provided our own implementation of PN in this experiment and future experiments. 
From Table \ref{tab:ablate}, it reveals that the addition of the relative features (R) 
has the most significant effect on the performance 
followed by the variance estimator (V) and the category-agnostic transformer (T). 
This is because relative features try to diminish the difference 
between feature dimensionality and the number of samples, 
and thus try to alleviate overfitting. 
On the other hand, PN+T has negligible improvement 
or slightly worse performance compared to the PN baseline. 
This is because prototypical networks tend to cluster 
same-class samples very close to one another and therefore
additional transformation stage (T) to map samples to prototype might be redundant. 
In certain cases, the complex non-linear transformation 
might over-fit to produce worse performance.  
It should be noted that higher ways in training 
do not always produce better performance. 
For example, in a 5-way 1-shot testing, 
PN+R produced a peak in performance for the 15-way training strategy 
with a dip in performance on either side. 
Similar pattern can be observed for the 5-way 5-shot testing results. 
The effect of relative features is also significant in case pairs of modules 
are added to the PN baseline. 
In Table \ref{tab:ablate}, 
we can see that PN+V+R and PN+R+T reached accuracy levels 
over 50\% and 70\% for the 5-way 1-shot 
and 5-way 5-shot testing cases, respectively,
but PN+V+T failed to do so. 
An interesting observation is that the combined effects of R+T 
mostly provided better performance than V+R even though 
V provided better performance than T. 
This suggests that adding modules upon the PN baseline 
did not always produce additive effects but they also produced interactive effects between the two modules.

\subsection{{Parameter Sensitivity Studies}}
\begin{figure}[]
\centering
\includegraphics[width=8cm]{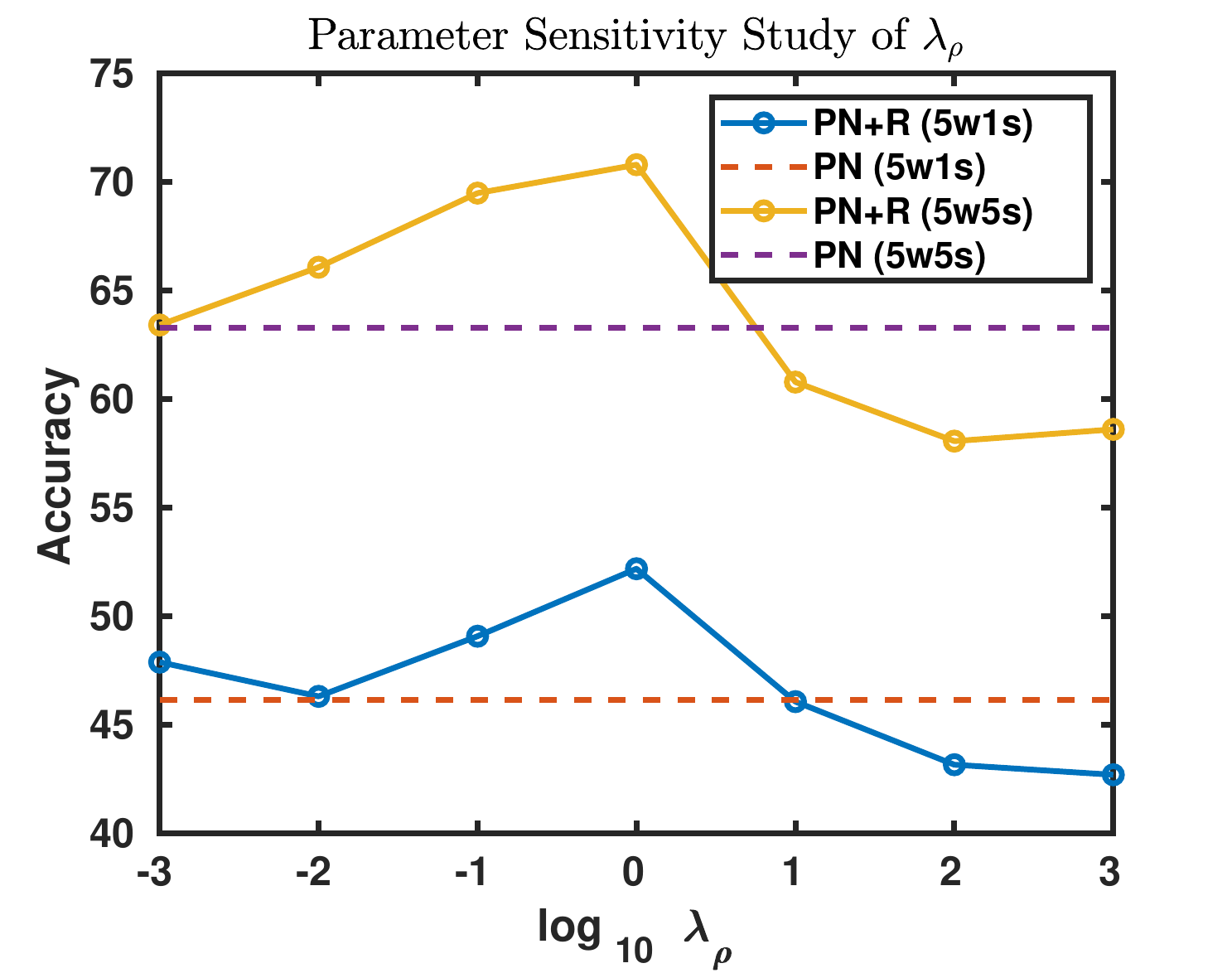}
\caption{Plot of accuracy with respect to $\lambda_{\rho}$ for 5-way 1-shot (5w1s) and 5-way 5-shot (5w5s) testing conditions with the prototypical network baseline. The dataset used is miniImagenet.}
\label{fig:lambdarho}
\end{figure}

{We also performed experiments} to find how 
the performance of PN+R varied with changing $\lambda_{\rho}$. 
The results are shown in Fig. \ref{fig:lambdarho} for both 5-way 1-shot 
and 5-way 5-shot testing conditions. The training condition for 5-way 1-shot testing is 20-way 1-shot and that for the 5-way 5-shot testing is 20-way 5-shot. 
The PN baseline is shown using the dotted line. 
From the plot, it is shown that the accuracy followed a bell-curve 
with the maximum accuracy observed at $\lambda_{\rho}=1$. 
It is recommended not to use $\lambda_{\rho} > 1$ 
as it caused degradation in performance, 
which was sometimes worse than the PN baseline. 
This is because putting excess weight on relative features diminishes 
the effect of absolute features that are crucial for recognition.

\begin{figure}[]
\centering
\includegraphics[width=8cm]{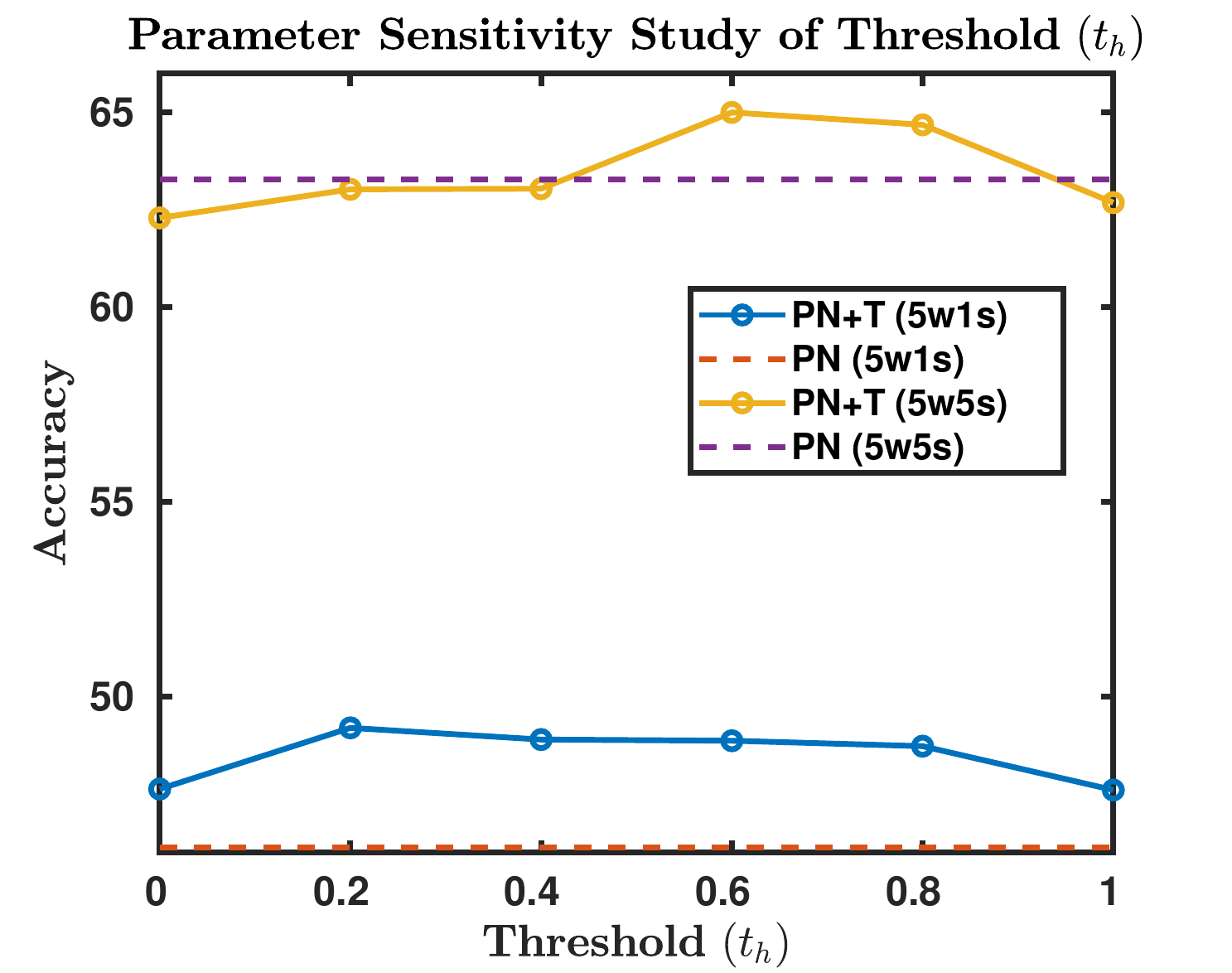}
\caption{{Plot of accuracy with respect to $t_h$ for 5-way 1-shot (5w1s) and 5-way 5-shot (5w5s) testing conditions with the prototypical network baseline. The dataset used is miniImagenet.}}
\label{fig:th}
\end{figure}

\begin{figure}[]
\centering
\includegraphics[width=8cm]{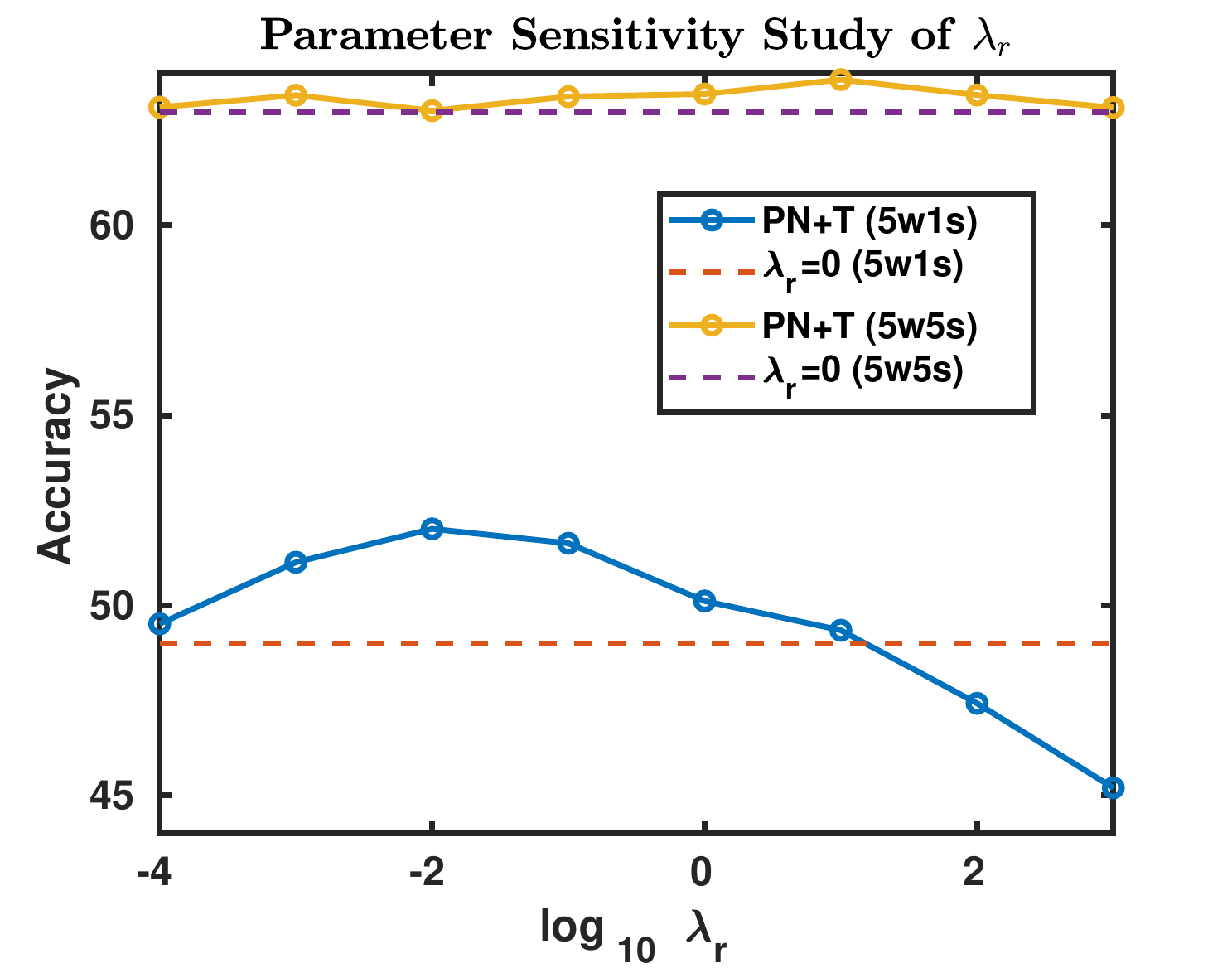}
\caption{{Plot of accuracy with respect to $\lambda_{r}$ 
for 5-way 1-shot (5w1s) and 5-way 5-shot (5w5s) testing conditions. 
The dataset used was miniImagenet.}}
\label{fig:lamr}
\end{figure}

{We also studied the effect of changing $t_h$ and $\lambda_r$ 
on the recognition performance for different testing shots. 
In Fig.~\ref{fig:th}, we see that the performance varied for different thresholds 
with a peak performance obtained for a value of $t_h$ between 0 and 1. 
In fact, for the higher shot configuration, 
the peak performance was obtained at a higher threshold. 
This is because for higher shots, 
the contribution of the few-shot sample mean was much more compared 
to the contribution of the base categories. 
As a result, a higher threshold $t_h$ was required 
to reduce the contribution of the base classes.}
{In Fig.~\ref{fig:lamr}, 
we observed how the recognition performance changed as $\lambda_r$ 
was varied for different shots.  
As expected, the peak performance was better than the baseline $\lambda_r=0$ 
shown in dashed lines. 
However, the sensitivity at the 5-shot configuration was less compared to 
that in the 1-shot configuration. 
{This is because}, for higher shots, 
the constraint corresponding to $\lambda_r$ - that the sample mean should be close to the prototype is automatically satisfied 
and therefore changing the value of $\lambda_r$ 
did not change the performance much.}

We did additional sensitivity studies of $t_h$ and $\lambda_r$ over a smaller range of values. 
The results are reported in Tables VI and VII 
for $t_h$ and $\lambda_r$, respectively. 
From the results, it showed that there was very little change 
when the parameters were varied over such a small range. 
However, the response was oscillatory probably 
because of the non-convexity of the loss functions used in our framework.

\begin{table}[]
\caption{Performance sensitivity with respect to threshold $t_h$ over a small range. The dataset used is miniImagenet.}
\label{tab:source}
\centering
\begin{tabular}{@{}cccccc@{}}
\toprule
$t_h$ & 0.02   & 0.04     & 0.06     & 0.08   & 0.1  \\ \midrule
5-way 1-shot   & 48.01 & 48.23 & 48.11 & 48.34 & 48.66 \\
5-way 5-shot   & 62.39 & 62.31 & 62.54 & 62.51 & 62.73 \\
\bottomrule
\end{tabular}
\end{table}

\begin{table}[]
\caption{Performance sensitivity with respect to $\lambda_r$ over a small range. The dataset used is miniImagenet.}
\label{tab:source}
\centering
\begin{tabular}{@{}cccccc@{}}
\toprule
$\lambda_r$ & 1e-4   & 2e-4    & 4e-4  &  8e-4  \\ \midrule
5-way 1-shot   & 49.51 & 50.64 & 50.50 & 50.78 \\
5-way 5-shot   & 63.11 & 63.26 & 63.36 & 63.34 \\
\bottomrule
\end{tabular}
\end{table}

\subsection{Feature Visualization}
\begin{figure}[]
\centering
\includegraphics[width=8cm]{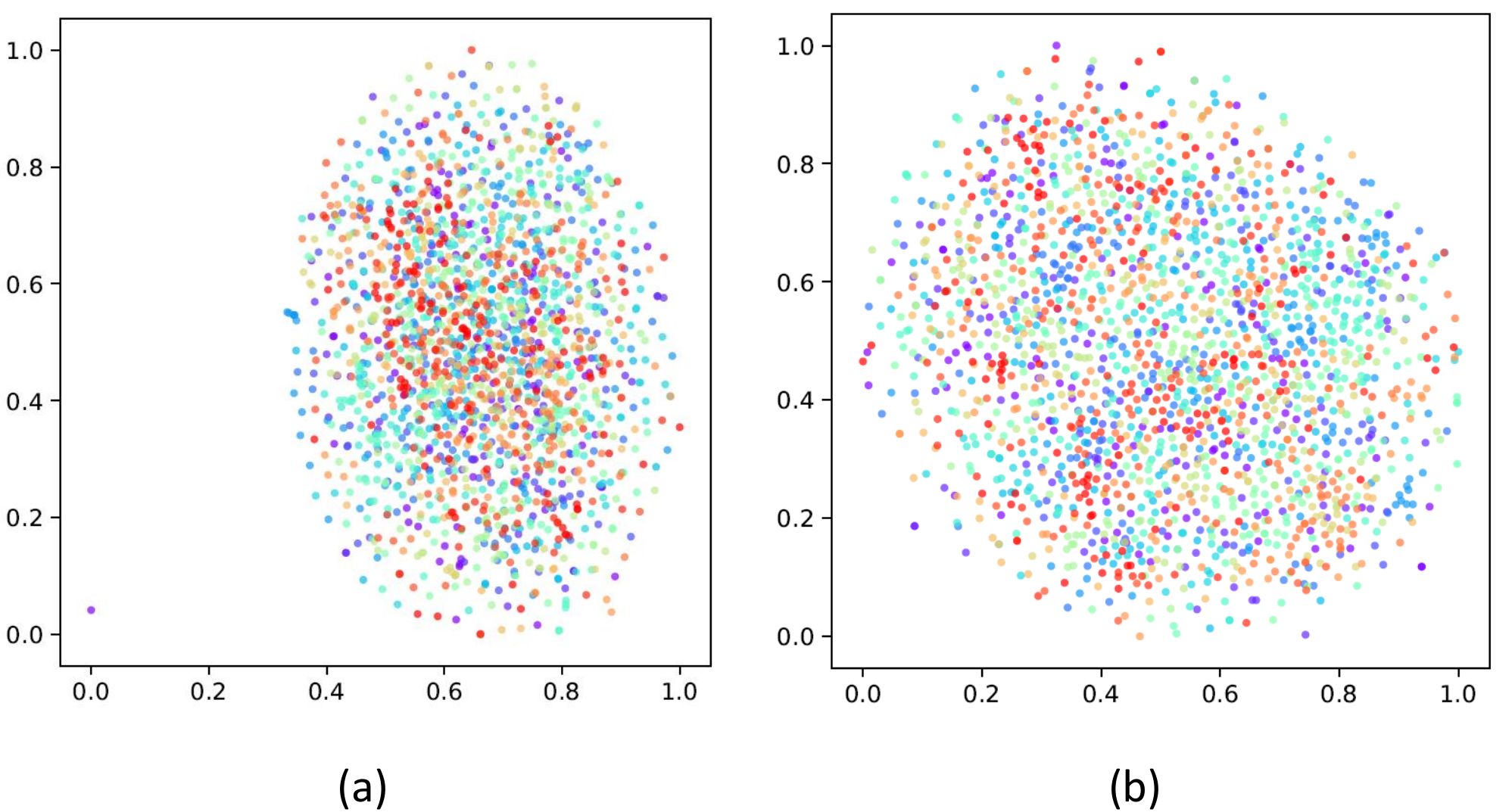}
\caption{t-SNE plot for (a) PN and (b) PN+R+V ($\lambda_{\rho}=1$). 
The dataset used was miniImagenet. 
Same color corresponds to different samples of the same category.}
\label{fig:tsne}
\end{figure}

We also visualized the features in two dimensions 
using t-SNE~\cite{maaten2008visualizing} as shown in Fig.~\ref{fig:tsne}. 
From Fig.~\ref{fig:tsne}(a), 
it is clear that PN produced a very compact feature space,  
where the classes were very difficult to distinguish. 
On the other hand, the features obtained using PN+R+V 
as shown in Fig.~\ref{fig:tsne}(b) were more distinguishable class-wise. 
This resulted in better recognition performance.

It is important to note that removing 
the outlier from Fig.~11(a) and rescaling the figure 
would make the image similar to Fig.~11(b). 
This is the point of difference between using Prototypical Network (PN) 
and our (PN+R+V) method. 
Using PN, we obtained more scaled-down features.
{Thus}, these features were closer to one another, 
resulting in more difficult classification 
compared to our (PN+R+V) method. 
However, distinguishing classes in both cases was complicated 
and that is why we used the Euclidean-distance-based differential nearest-neighbor classifier.

\subsection{Convergence Results}
\begin{figure}[]
\centering
\includegraphics[width=8cm]{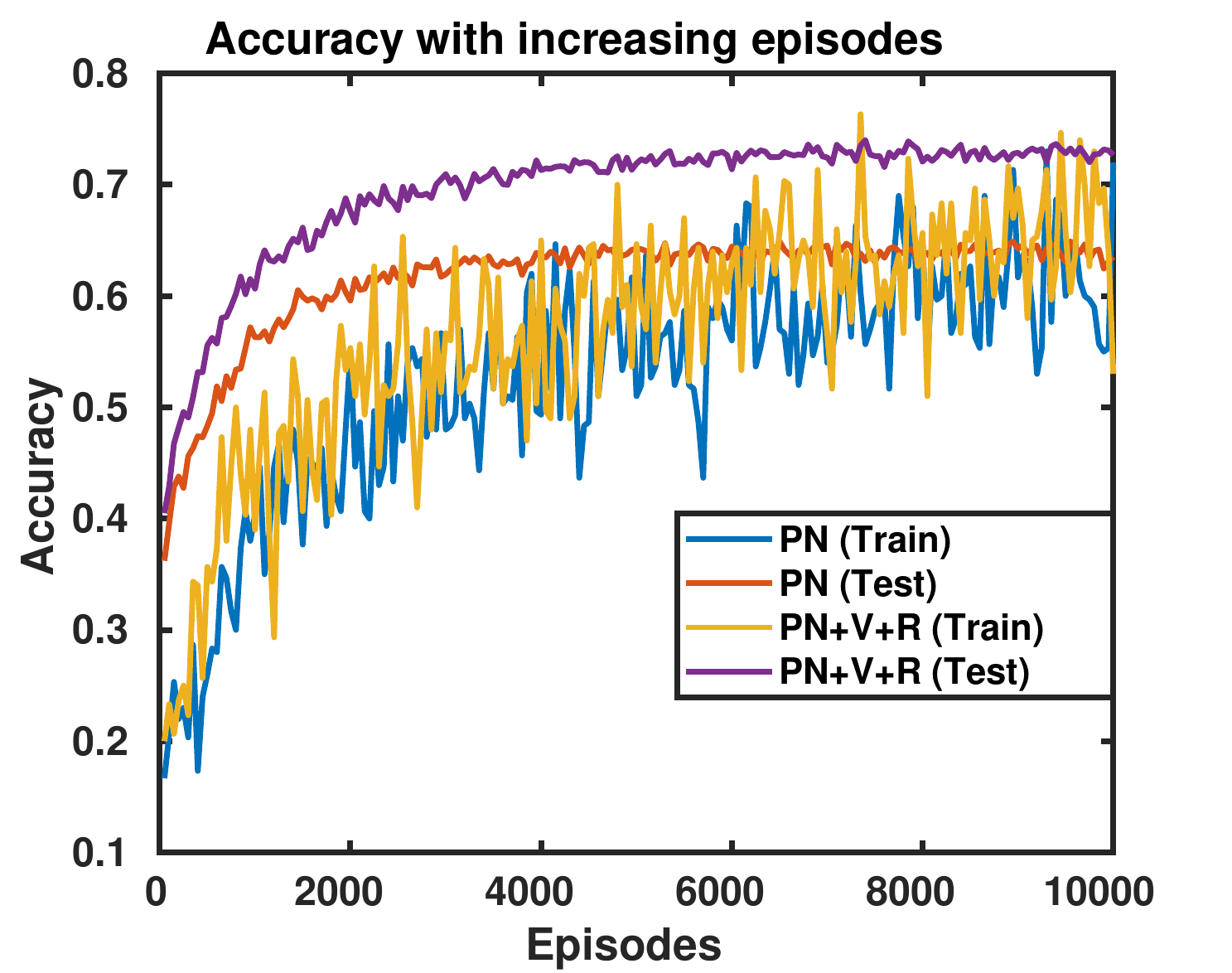}
\caption{Training and test accuracy with increasing number of episodes 
for the prototypical network (PN) baseline 
and our proposed approach using relative features and variance estimator (PN+V+R).}
\label{fig:episode}
\end{figure}

We also reported the training and testing performance 
with increasing training episodes in Fig.~\ref{fig:episode}. 
We used the 5-way 5-shot and 20-way 5-shot settings for testing and training, respectively. 
As shown in Fig.~\ref{fig:episode}, 
the test accuracy for PN+V+R rose fast compared to that of PN. 
Also, the training accuracy was quite noisy. 
This is because each training episode produced a newer set of categories 
and therefore there was a high variance in the training accuracy.  

\subsection{Effect of Number of Samples}
\begin{figure}[]
\centering
\includegraphics[width=8cm]{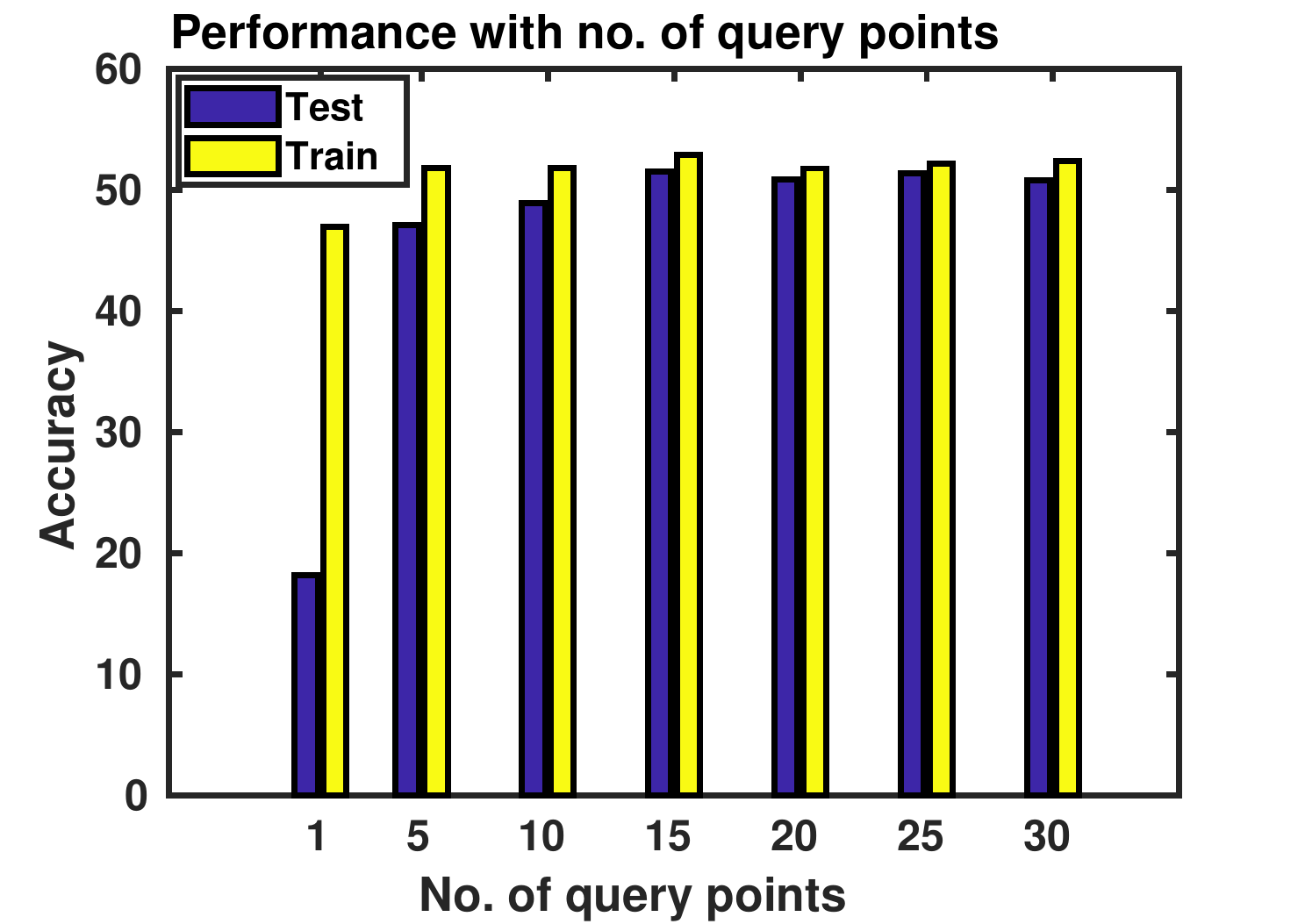}
\caption{Plot of accuracy when the number of training query points is fixed and the number of test query points is varied and vice-versa. 
The dataset used was miniImagenet.}
\label{fig:transduct}
\end{figure}

Since the relative features are constructed using both the support and query points, 
it is worthwhile to note the effect on recognition performance 
by changing the number of query points per class in the training and testing stages. 
We performed two experiments for the PN+R case. 
The first experiment considered the situation 
when the number of training query points per class was fixed at 15 
and the number of test query points was varied. 
The second experiment considered the situation 
when the number of test query points per class was fixed at 15 
and the number of training points was varied. 
In Fig.~\ref{fig:transduct}, 
it is shown that as the number of query points increased, 
the recognition performance increased and it became saturated after a while. 
This is because query points beyond a certain quantity 
did not provide additional second-order structural information. 
Also, from the poor performance in the case of one test query sample, 
it is evident that having sufficient query samples 
in the testing stage was more important than 
having sufficient quantity of query samples in the training stage.

\subsection{Effect of Base Categories}
We also evaluated how the performance of PN+T varied 
as the number of source base categories changed. 
Results are shown in Table \ref{tab:source}. 
The recognition performance increased 
with the increasing number of source categories. 
This is because the increasing number of source categories 
trained a robust feature space. 
Also, the probability of finding relevant categories 
became more for the category-agnostic transformation stage. 
{The performance} of the category-agnostic transformation became poorer 
at higher shots compared to PN. This is because the transformation 
became closer to identity and its significance became less.

Till now, we have tested our proposed approach on the novel categories. 
It is also important to test our proposed approach on base categories 
since they are more common and are likely to be observed 
more frequently compared to novel categories. 
The results of applying our proposed approach to the base categories 
are shown in Table \ref{tab:baseclass} for different testing settings. 
As expected, the performance on base categories was better compared to 
that of novel categories. 
Furthermore, our proposed approach (PN+V+R) 
produced better results as compared to PN.

\begin{table}[]
\caption{Performance analysis as the number of base categories is varied for the PN+T case. The dataset used is miniImagenet.}
\label{tab:source}
\centering
\begin{tabular}{@{}lccccc@{}}
\toprule
Source No. & 20    & 30     & 40     & 50   & 60  \\ \midrule
5-way 1-shot (PN)   & 40.10 & 42.196 & 44.14 & 45.66 & 45.74 \\
5-way 1-shot (PN+T)  & 41.61 & 43.74 & 45.48 & 46.82 & 47.20 \\
5-way 5-shot (PN)   & 45.89 & 51.93 & 55.95 & 59.796 & 60.96 
\\
5-way 5-shot (PN+T)   & 43.89 & 50.93 & 55.85 & 59.70 & 61.14 
\\
\bottomrule
\end{tabular}
\end{table}

\begin{table}[]
\caption{Performance comparison of testing on the base training classes.}
\label{tab:baseclass}
\centering
\begin{tabular}{cccc}
\hline
\begin{tabular}[c]{@{}c@{}}5-way 5-shot \\ (PN)\end{tabular} & \begin{tabular}[c]{@{}c@{}}5-way 1-shot\\  (PN)\end{tabular} & \begin{tabular}[c]{@{}c@{}}5-way 5-shot \\ (PN+V+R)\end{tabular} & \begin{tabular}[c]{@{}c@{}}5-way 1-shot \\ (PN+V+R)\end{tabular} \\ \hline
82.236                                                       & 58.409                                                       & 85.293                                                           & 64.111                                                           \\ \hline
\end{tabular}
\end{table}

\subsection{Analysis of Category-agnostic Transformation}
We also carried out the ablation analysis of PN+T; 
that is, the addition of the category-agnostic transformer (T) 
on top of the prototypical network baseline (PN).  
As described previously, the category-agnostic transformer (T) 
consists of three modules - the neural-network-based transformer ($\text{T}_{11}$), 
the residual connection ($\text{T}_{12}$), 
and the contribution of the base prototypes ($\text{T}_{2}$). 
From Table \ref{tab:transform}, 
we can see that the addition of these modules gradually 
improved the recognition performance, 
suggesting that the addition of all these modules was important. 
The method PN+$\text{T}_{11}$+$\text{T}_{12}$+$\text{T}_2$ 
used a threshold $t_h=0.02$. 
It is important to note that using PN+$\text{T}_{11}$+$\text{T}_{12}$ 
was equivalent to PN+$\text{T}_{11}$+$\text{T}_{12}$+$\text{T}_2$ 
with threshold $t_h=1$. 
We also performed an additional experiment 
using the method PN+$\text{T}_{11}$+$\text{T}_{12}$+$\text{T}_2$ 
with threshold $t_h=0$. 
Using $t_h=0$, we obtained an accuracy of $47.63\%$ and $62.29\%$ 
on the 5-way 1-shot and 5-way 5-shot classification tasks, respectively. 
The recognition performance was worse compared to using $t_h=0.02$ 
because $t_h=0$ caused all the base classes and therefore irrelevant classes 
to contribute to the category-agnostic transformation thus causing a negative transfer. 

\begin{table}[]
\caption{{Ablation analysis of each component} of the category-agnostic transformer. The dataset used was miniImagenet.}
\label{tab:transform}
\centering
\begin{tabular}{@{}lll@{}}
\toprule
Method             & 5-way 1-shot & 5-way 5-shot \\ \midrule
PN+$\text{T}_{11}$           & 47.393       & 62.411       \\
PN+$\text{T}_{11}$+$\text{T}_{12}$      & 48.604       & 62.683    \\
PN+$\text{T}_{11}$+$\text{T}_{12}$+$\text{T}_2$ & 49.002       & 63.024       \\ \bottomrule
\end{tabular}
\end{table}

The category agnostic transformer consisted of 
contribution of the base categories 
as described mathematically through $\mathbf{f}_{\text{T}_2}$. 
Using the threshold mechanism, 
only relevant base categories were selected for contribution 
because these categories were closer to the novel category 
in the feature space compared to the irrelevant base categories. 
Using the thresholded probability vector $\mathbf{p}_c^{th}$, 
we selected the top three relevant base categories for a few novel categories. 
The results are shown in Table \ref{tab:relevant}. 
As an example, all the top relevant categories 
for the African hunting dog have canine features. 
The relevant categories for the mixing bowl seem to fit in context. 
Pictures of Consomme and Hotdog are generally shown in plates or bowls. 
Also, the relevant categories of nematode, 
a worm-like organism involved insects and snakes. 
There could be erroneous selections like harvestman spider 
being the most relevant category for the Golden-retriever dog. 
This suggested that an additional class relevance criterion 
based on WordNet~\cite{miller1995wordnet} might be more appropriate.

\begin{table}[]
\caption{Novel categories and top three relevant base categories.}
\label{tab:relevant}
\centering
\begin{tabular}{@{}cccc@{}}
\toprule
\textbf{Novel /Relevant Class} & \textbf{Rank 1} & \textbf{Rank 2}  & \textbf{Rank 3} \\ \midrule
African hunting dog                 & Saluki          & Arctic Fox       & Komondor        \\
Mixing bowl                         & Consomme        & Hotdog           & Ear             \\
Golden-retriever                    & Harvestman      & Miniature poodle & Bolete          \\
Nematode                            & Green-mamba     & Lady-bug         & Spider-web      \\ \bottomrule
\end{tabular}
\end{table}

\section{Conclusions}
We have proposed a two-stage framework for few-shot learning of image recognition. 
The framework has contributions at both the feature extraction stage 
and the classification stage of image recognition. 
At the feature extraction stage, we proposed the use of relative-feature representation 
as well as the Mahalanobis distance metric with predictable variance. 
For the classification stage, 
we proposed a category-agnostic transformation 
that produces class prototypes from class samples. 
Results on standard few-shot learning datasets showed our approach 
to be comparable or even better than previous approaches. 
We also provided further analysis on our model 
and concluded that the relative-feature component 
was mostly responsible for the improvement of the performance 
of our proposed approach. 
In the future, we would like to extend our work to zero-shot classification, 
where we do not have any support samples from the novel class 
but only high-level semantic information for each of these classes.


%

%

%

\ifCLASSOPTIONcaptionsoff
  \newpage
\fi



%
\bibliographystyle{IEEEbib}
\bibliography{final}

%
%

%

\begin{IEEEbiography}
[{\includegraphics[width=1in,height=1.25in,clip,keepaspectratio]{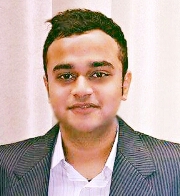}}]{Debasmit Das} 
received the B.Tech. degree in Electrical Engineering from the Indian Institute of Technology, Roorkee in 2014. He is currently pursuing his Ph.D. degree at the School of Electrical and Computer Engineering, Purdue University. His research interest is in label-efficient learning especially in transfer learning problems like domain adaptation, zero-shot learning etc. He is a student member of IEEE and he also regularly reviews machine learning articles for IEEE, ACM and Springer journals.
\end{IEEEbiography}

\begin{IEEEbiography}
[{\includegraphics[width=1in,height=1.25in,clip,keepaspectratio]{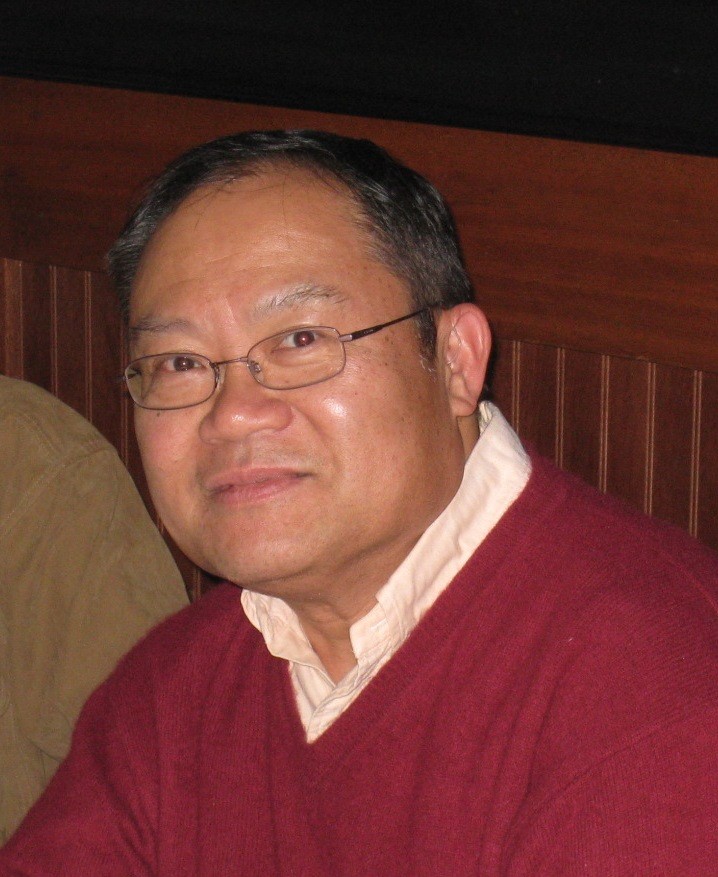}}]{C.S. George Lee}
is a Professor of Electrical and Computer Engineering at Purdue University, West Lafayette, Indiana.  
His current research focuses on transfer learning and skill learning, human-centered robotics, and neuro-fuzzy systems.  He has published extensively in those areas with over 250 archival publications, co-authored two graduate textbooks, \emph{Robotics: Control, Sensing, Vision, and Intelligence} (McGraw-Hill, 1986) and \emph{Neural Fuzzy Systems: A Neuro-Fuzzy Synergism to Intelligent Systems} (Prentice-Hall, 1996), and over 20 book chapters.  Dr. Lee is an IEEE Fellow, a recipient of the IEEE Third Millennium Medal Award, the Saridis Leadership Award and the Distinguished Service Award from the IEEE Robotics and Automation Society.  Dr. Lee received his Ph.D. degree from Purdue University, West Lafayette, Indiana, and his M.S.E.E. and B.S.E.E. degrees from Washington State University, Pullman, Washington.  More detailed information can be obtained from his website at https:/engineering.purdue.edu/artlab/
\end{IEEEbiography}






\end{document}